\documentclass{new_tlp}
\usepackage{mathptmx}

\usepackage[draft]{hyperref}
\usepackage[utf8]{inputenc}

\usepackage{times}
\usepackage{pifont}
\usepackage{textcomp}
\usepackage{url}
\usepackage{booktabs}
\usepackage[small]{caption}
\usepackage{float}
\usepackage{verbatim}
\usepackage{varwidth}
\usepackage{pifont}
\usepackage{url}
\usepackage[misc]{ifsym}
\usepackage{graphicx}
\usepackage{booktabs}
\usepackage{epstopdf}
\usepackage{longtable}
\usepackage{pgfpages}
\usepackage{multirow}
\usepackage{url}
\usepackage{multicol}

\usepackage{subfig}

\usepackage{amsmath} 
\usepackage{algorithm}
\usepackage[noend]{algpseudocode}
\algnotext{EndFor}
\algnotext{EndIf}

\DeclareMathOperator*{\argmax}{arg\,max}

\usepackage{booktabs,array,tabularx} 
\makeatletter
\let\@tabclassz =\@classz
\let\@tabclassiv =\@classiv
\makeatother
\newcolumntype{C}{>{\centering\arraybackslash}X}

\newtheorem{example}{Example}

\newcommand\bcmdtab{\noindent\bgroup\tabcolsep=0pt%
  \begin{tabular}{@{}p{10pc}@{}p{20pc}@{}}}
\newcommand\ecmdtab{\end{tabular}\egroup}

\newcommand{\nbfnew}{\textsf{\scriptsize not}}

\newcommand{\weak}{:\text{\texttildelow} \ }

\newcommand{\myhappens}{\textsf{\scriptsize happensAt}}
\newcommand{\myholdsAt}{\textsf{\scriptsize holdsAt}}
\newcommand{\myinitiatedAt}{\textsf{\scriptsize initiatedAt}}
\newcommand{\myterminatedAt}{\textsf{\scriptsize terminatedAt}}
\newcommand{\mynbf}{\textsf{\scriptsize not}}
\newcommand{\myhappenss}{\textsf{\footnotesize happensAt}}
\newcommand{\myholdsAts}{\textsf{\footnotesize holdsAt}}
\newcommand{\myinitiatedAts}{\textsf{\footnotesize initiatedAt}}
\newcommand{\myterminatedAts}{\textsf{\footnotesize terminatedAt}}
\newcommand{\mynbfs}{\textsf{\footnotesize not}}

\newcommand{\xhail}{\textsf{\footnotesize XHAIL}}
\newcommand{\maxmargin}{\textsf{\footnotesize MaxMargin}}
\newcommand{\tal}{\textsf{\footnotesize TAL}}
\newcommand{\oled}{\textsf{\footnotesize OLED}}
\newcommand{\ilasp}{\textsf{\footnotesize ILASP}}
\newcommand{\ilaspfour}{\textsf{\footnotesize ILASP4}}
\newcommand{\iled}{\textsf{\footnotesize ILED}}
\newcommand{\iledhc}{\textsf{\footnotesize ILED-HC}}
\newcommand{\woled}{\textsf{\footnotesize WOLED}}
\newcommand{\woledasp}{\textsf{\footnotesize WOLED-ASP}}
\newcommand{\woledmln}{\textsf{\footnotesize WOLED-MLN}}
\newcommand{\lomrf}{\textsf{\footnotesize LoMRF}}
\newcommand{\lpsolve}{\textsf{\footnotesize lpsolve}}
\newcommand{\hand}{\textsf{\footnotesize HandCrafted}}

\newcommand{\adagrad}{\textsf{\footnotesize AdaGrad}}

\newcommand{\ec}{\textsf{\footnotesize EC}}
\newcommand{\clingo}{\textsf{\footnotesize Clingo}}

\title[Theory and Practice of Logic Programming]
      {Online Learning Probabilistic Event 
      Calculus Theories in Answer Set Programming}

\author[N. Katzouris, A. Artikis and G. Paliouras]{
NIKOS KATZOURIS$^{1}$, ALEXANDER ARTIKIS$^{2,1}$ and GEORGIOS PALIOURAS$^{1}$,  \\
$^1$Institute of Informatics \& Telecommunications,\\ National Center for Scientific Research (NCSR) ``Demokritos'', Athens, Greece\\
$^2$Department of Maritime Studies, University of Piraeus, Piraeus, Greece\\
\email{\{nkatz,a.artikis,paliourg\}@iit.demokritos.gr}
}

\begin{document}

\label{firstpage}

\maketitle

  \begin{abstract}
    Complex Event Recognition (CER) systems detect event occurrences in streaming time-stamped input using predefined event patterns. Logic-based approaches are of special interest in CER, since, via Statistical Relational AI, they combine uncertainty-resilient reasoning with time and change, with machine learning, thus alleviating the cost of manual event pattern authoring. We present a system based on Answer Set Programming (ASP), capable of probabilistic reasoning with complex event patterns in the form of weighted rules in the Event Calculus, whose structure and weights are learnt online. We compare our ASP-based implementation with a Markov Logic-based one and with a number of state-of-the-art batch learning algorithms on CER datasets for activity recognition, maritime surveillance and fleet management. Our results demonstrate the superiority of our novel approach, both in terms of efficiency and predictive performance. This paper is under consideration for publication in Theory and Practice of Logic Programming (TPLP).
  \end{abstract}



\section{Introduction}
\label{sec:intro}

Complex Event Recognition (CER) systems \cite{cugola2012processing} detect occurrences of \emph{complex events} (CEs) in streaming input, defined as spatio-temporal combinations of \emph{simple events} (e.g. sensor data), using a set of CE patterns. Since such patterns are not always known beforehand, machine learning algorithms for discovering them from data are highly useful. Thanks to their efficiency, online learning algorithms are of special interest. Such algorithms should be resilient to noise \& uncertainty, which are ubiquitous in temporal data streams \cite{DBLP:journals/corr/AlevizosSAP17}, while taking into account commonsense phenomena \cite{mueller2014commonsense}, which often characterize dynamic application domains, such as CER.  

Logic-based CER systems \cite{artikis2012logic}  
stand up to these challenges. They combine reasoning under uncertainty with machine learning, via Statistical Relational AI techniques \cite{raedt2016statistical}, while supporting reasoning with time and change, via action formalisms such as the Event Calculus \cite{artikis2015event}.  

We advance the state of the art in online learning for CER by proposing \woled \ (Online Learning of Weighted Event Definitions), an algorithm that learns CE patterns in the form of weighted rules in the Event Calculus. 
The proposed algorithm is based entirely on Answer Set Programming (ASP) \cite{lifschitz2019answer}, which allows to take advantage of the grounding, solving, optimization and uncertainty modeling abilities of modern answer set solvers, while employing structure learning techniques from non-monotonic Inductive Logic Programming (ILP) \cite{de2008logical}, which are easily implemented in ASP.

We compare \woled's ASP-based implementation to an MLN-based one, 
and to a number of state of the art online \& batch structure \& weight learning algorithms, on three CER datasets for \emph{activity recognition}, \emph{maritime surveillance} and \emph{vehicle fleet management}. Our results demonstrate the superiority of \woled, both in terms of efficiency and predictive performance.

This paper is an extended version of \cite{kr2020}, which has been nominated as a candidate for TPLP's rapid publication track by KR2020's program committee.

\section{Related Work}
\label{sec:related}

Event Calculus-based CER \cite{artikis2015event} was combined with MLNs in \cite{skarlatidis2015probabilistic}, in order to deal with the noise and uncertainty of CER applications. An inherent limitation of this approach is the fact that the non-monotonic semantics of the Event Calculus is incompatible with the open-world semantics of MLNs. Therefore, performing inference with Event Calculus-based MLN theories calls for extra, costly operations, such as computing the completion of a theory \cite{mueller2014commonsense}, in order to endow MLNs' first-order logic representations with a non-monotonic semantics. We bridge this gap via translating probabilistic inference with MLNs into an optimization task in ASP, which naturally supports non-monotonic and commonsense reasoning. This also allows to delegate probabilistic temporal reasoning and machine learning tasks to sophisticated, off-the-shelf answer set solvers.

Translating MLN inference in ASP has been put forth in \cite{DBLP:conf/kr/LeeW16,DBLP:journals/tplp/LeeTW17}. This line of work is mostly concerned with theoretical aspects of the translation, limiting applications to simple, proof-of-concept examples. Although we do rely on the theoretical foundation of this work, we take a more application-oriented stand-point and investigate the usefulness of these ideas in challenging domains, such as CER. We also propose a methodology for online structure and the weight learning using ASP tools.

Regarding machine learning, a number of algorithms in non-monotonic Inductive Logic Programming (ILP), such as \xhail \ \cite{ray2009nonmonotonic}, \tal \ \cite{athakravi2013learning} and \ilasp \ \cite{ilasp} are capable of learning Event Calculus theories, see \cite{katzourisPhd} for a comprehensive review of such approaches. These algorithms are batch learners, they are thus poor matches to the online nature of CER applications. Moreover, they learn crisp logical theories, thus their ability to cope with noise and uncertainty is limited. Existing online learning algorithms are either crisp learners \cite{DBLP:journals/fgcs/KatzourisAP19}, or they rely on MLNs \cite{DBLP:conf/pkdd/KatzourisMAP18,michelioudakis2016mathtt}, so they suffer from the same limitations discussed earlier in this section. A recent online learner based on probabilistic theory revision \cite{DBLP:journals/ml/GuimaraesPZ19} is limited to Horn logic and cannot handle Event Calculus reasoning.      

\section{Background}
\label{sec:background}

\textbf{Answer Set Programming.} In what follows a rule $r$ is an expression of the form $\alpha \leftarrow \delta_1,\ldots,\delta_n$, where $\alpha$ is an atom, called the head of $r$, 
$\delta_i's$ are literals (possibly negated atoms), which collectively form the body of $r$ 
and commas in the bodies of rules denote conjunction. 
A rule is ground if it contains no variables and a grounding of a rule $r$ is called an instance of $r$. A (Herbrand) interpretation is a collection of true ground facts. An interpretation $I$ satisfies an atom $\alpha$ iff $\alpha \in I$. $I$ satisfies a ground rule iff satisfying each literal in the body implies that the head atom is also satisfied and it satisfies a non-ground rule $r$ if it satisfies all ground instances of $r$. 
An interpretation $I$ is a model of a logic program $\Pi$ (collection of rules) if it satisfies every rule in $\Pi$ and it is a minimal model if no strict subset of $I$ has this property. An interpretation $I$ is an answer set of $\Pi$ iff it is a minimal model of the reduct of $\Pi$, i.e. the program that results by removing all rules with a negated body literal not satisfied by $I$ and removing all negated literals from the bodies of the remaining rules.  

A choice rule is an expression of the form $\{\alpha\} \leftarrow \delta_1,\ldots,\delta_n$ with the intuitive meaning that whenever the body $\delta_1,\ldots,\delta_n$ is satisfied by an answer set $I$ of a program that includes the choice rule, instances of the head $\alpha$ are arbitrarily included in $I$ (satisfied) as well. A weak constraint is an expression of the form $\weak \delta_1,\ldots,\delta_n. [w]$, where $\delta_i$'s are literals and $w$ is an integer. The intuitive meaning of a weak constraint $c$ is that the satisfaction of the conjunction $\delta_1,\ldots,\delta_n$ by an answer set $I$ of a program that includes $c$ incurs a cost of $w$ for $I$. Inclusion of weak constraints in a program triggers an optimization process that yields answer sets of minimum cost. 

\textbf{The Event Calculus} is a temporal logic for reasoning about events and their effects. Its ontology comprises time points (integers), fluents, i.e. properties which have certain values in time, and events, i.e. occurrences in time that may affect fluents and alter their value. Its axioms incorporate the commonsense law of inertia, according to which fluents persist over time, unless they are affected by an event. Its basic predicates and axioms are presented in Table \ref{table:stream1}(a), (b). Axiom (1) states that a fluent $F$ holds at time $T$ if it has been initiated at the previous time point, while Axiom (2) states that $F$ continues to hold unless it is terminated. Definitions of \myinitiatedAts/2 and \myterminatedAts/2 predicates are provided in a application-specific manner. 

Using the Event Calculus in a CER context allows to reason with CEs that have duration in time and are subject to commonsense phenomena, via associating CEs to fluents. In this case, a set of CE patterns is a set of 
\myinitiatedAts/2 and \myterminatedAts/2 rules.

\begin{table*}[t]
\scriptsize
\begin{minipage}{\textwidth}

\begin{tabular}{ p{0.2\textwidth} p{0.8\textwidth}} 
\toprule
\textbf{\emph{(a) }} & ~\\
\textbf{\emph{Predicate:}} & \textbf{\emph{Meaning:}}\\
$\myhappens(E,T)$ &  Event $E$ occurs at time $T$.\\
$\myinitiatedAt(F,T)$ &  At time $T$, a period of time for which fluent $F$ holds is initiated.\\
$\myterminatedAt(F,T)$ &  At time $T$, a period of time for which fluent $F$ holds is terminated.\\
$\myholdsAt(F,T)$ &  Fluent $F$ holds at time $T$.\\
\midrule
\end{tabular}\\

\begin{tabular}{ p{\textwidth} }
\vspace*{-0.7cm}
\begin{multline*}
\begin{array}{l}
\hspace*{-0.5cm}\textbf{\emph{(b) Axioms of the Event Calculus}}\\
\myholdsAt(F,T+1) \leftarrow \myinitiatedAt(F,T). \hspace*{2.6cm} (1)\\
\myholdsAt(F,T+1) \leftarrow \myholdsAt(F,T), \mynbf \ \myterminatedAt(F,T).\hspace*{0.36cm} (2)
\end{array}
\end{multline*}\\
\end{tabular}

\vspace*{-0.5cm}

\begin{tabular}{ p{0.5\textwidth} p{0.5\textwidth}} 
\hline
\textbf{\emph{(c) }} & ~\\
\vspace*{-0.7cm}
\begin{multline*}
\begin{array}{l}
\hspace*{-0.5cm}\textbf{\emph{Observations $I_1$ at time 1:}}\\
\hspace*{-0.5cm}\{\myhappens(walk(id_1),1),\myhappens(walk(id_2),1) \\
\hspace*{-0.5cm}\quad coords(id_1,201,454,1),coords(id_2,230,440,1),\\
\hspace*{-0.5cm}\quad direction(id_1,270,1),direction(id_2,270,1)
\}\\
\end{array}
\end{multline*}
&
\vspace*{-0.7cm}
\begin{multline*}
\begin{array}{l}
\textbf{\emph{Target CE instances at time 1:}}\\
\{
\myholdsAt(move(id_1,id_2),2),
\quad \myholdsAt(move(id_2,id_1),2)
\}
\end{array}
\end{multline*}
\end{tabular}\\

\begin{tabular}{ p{\textwidth} }
\vspace*{-1cm}
\begin{multline*}
\begin{array}{l}
\hspace*{-0.5cm}\textbf{\emph{(d) Weighted CE patterns:}}\\
\hspace*{-0.5cm}1.283 \ \myinitiatedAt(move(X,Y),T) \leftarrow \myhappens(walk(X),T),\myhappens(walk(Y),T),  
close(X,Y,25,T),orientation(X,Y,45,T). \\
\hspace*{-0.5cm}0.923 \ \myterminatedAt(move(X,Y),T) \leftarrow \myhappens(inactive(X),T), \mynbf \ close(X,Y,30,T). 
\vspace*{-0.5cm}
\end{array}
\end{multline*}\\
\hline
\end{tabular}
\end{minipage}
\vspace*{-0.3cm}
\caption{\textbf{(a), (b)} The basic predicates and the Event Calculus axioms. \textbf{(c)} Example CAVIAR data. At time point 1 person with $\mathit{id_1}$ is \emph{walking}, her $(X,Y)$ coordinates are $(201,454)$ and her direction is $270^{\circ}$. The target CE atoms (true state -- supervision) for time point 1 state that persons $id_1$ and $id_2$ are moving together at the next time point. \textbf{(d)} An example of two domain-specific axioms in the \ec. E.g. the first rule dictates that \emph{moving together} between two persons $X$ and $Y$ is initiated at time $T$ if both $X$ and $Y$ are walking at time $T$, their euclidean distance is less than 25 pixel positions and their difference in direction is less than $45^{\circ}$. The second rule dictates that \emph{moving together} between $X$ and $Y$ is terminated at time $T$ if one of them is standing still at time $T$ and their euclidean distance at $T$ is greater that 30.}
\label{table:stream1}
\end{table*}
\normalsize

As an example we use the task of activity recognition, as defined in the CAVIAR project\footnote{\url{http://homepages.inf.ed.ac.uk/rbf/CAVIARDATA1/}}. The CAVIAR dataset consists of videos of a public space, where actors perform some activities. These videos have been manually annotated by the CAVIAR team to provide the ground truth for two types of activity. The first type, corresponding to simple events, consists of knowledge about a person's activities at a certain video frame/time point (e.g. \emph{walking, standing still} and so on). The second type, corresponding to CEs/fluents, consists of activities that involve more than one person, for instance two people \emph{moving together, meeting each other} and so on. The aim is to detect CEs as of combinations of simple events and additional domain knowledge, such as a person's position and direction.

Table \ref{table:stream1}(c) presents an example of CAVIAR data, consisting of \emph{observations} for a particular time point, in the form of an interpretation $I_1$. A stream of interpretations is matched against a set of CE patterns (initiation/termination rules -- see Table \ref{table:stream1}(d)), to infer the truth values of CE instances in time, using the Event Calculus axioms as a reasoning engine. We henceforth call the atoms corresponding to CE instances whose truth values are to be inferred/predicted, \emph{target CE instances}. Table \ref{table:stream1}(c) presents the target CE instances corresponding to the observations in $I_1$. 

In what follows the CE patterns included in a logic program $\Pi$ are associated with real-valued weights, defining a probability distribution over answer sets of $\Pi$. Similarly to Markov Logic, where a possible world may satisfy a subset of the formulae in an MLN, and the weights of the formulae in a unique, maximal such subset determine the probability of the possible world, an answer set of a program with weighted rules may satisfy subsets of these rules, and these rules' weights determine the answer set's probability. Based on this observation, \cite{DBLP:conf/kr/LeeW16} propose to assign probabilities to answer sets of a program $\Pi$ with weighted rules as follows: For each interpretation $I$, first find the maximal subset $R_I$ of the weighted rules in $\Pi$ that are satisfied by $I$. Then, assign to $I$ a weight $W_{\Pi}(I)$ proportional to the sum of weights of the rules in $R_I$, if $I$ is an answer set of $R_I$, else assign zero weight. Finally, define a probability distribution over answer sets of $\Pi$ by normalizing these weights. 

Formally, let $w_r$ be the weight of rule $r$ and $\mathsf{ans}(\Pi)$ the set of all interpretations $I$ which are answer sets of $R_I$ and which, moreover, satisfy all hard-constrained rules in $\Pi$ (rules without weights). Then:

\small
\begin{multicols}{2}
  \begin{equation}
    \label{eq:weight}
W_{\Pi}(I)=\left\{\begin{array}{ll}
\exp \left(\sum\limits_{r \in R_I} w_r\right) & \text { if } I \in \mathsf{ans}(\Pi) \\
0 & \text { otherwise }
\end{array}\right.
  \end{equation}\break
  \begin{equation}
    \label{eq:probability}
P_{\Pi}(I)=\frac{W_{\Pi}(I)}{\sum\limits_{J \in \mathsf{ans}(\Pi)} W_{\Pi}(J)}
  \end{equation}
\end{multicols}
\normalsize

\section{Structure \& Weight Learning in ASP}
\label{sec:woled}

The task that \woled, our proposed algorithm, addresses is to online learn the structure and weights of CE patterns, while using their current version at each point in time to perform CER in the streaming input. We adopt a standard online learning approach consisting of the following steps: at time $t$ the learner maintains a theory $H_t$  (weighted CE pattern set, as in Table \ref{table:stream1}(c)), has access to some static background knowledge (e.g. the axioms of the Event Calculus -- Table \ref{table:stream1}(a)) and receives an interpretation $I_t$, consisting of a data mini-batch (as in Table \ref{table:stream1}(b)). Then (i) the learner performs inference (CER) with $B\cup H_t$ on $I_t$ ($B$ is the background knowledge) and generates a ``predicted state'', consisting of inferred \myholdsAt/2 instances of the target predicate. Via closed-world assumption, all such instances not present in the predicted state are assumed false; (ii) if available, the true state, consisting of the actual truth values of the predicted atoms is revealed; (iii) the learner identifies erroneous predictions via comparing the predicted state to the true one, and uses these mistakes to update the structure and the weights of the CE patterns in $H_t$, yielding a new theory $H_{t+1}$. We next discuss each of these steps.

\subsection{Generating the Inferred State}

To make predictions with the weighted CE patterns in the incoming data interpretations, \woled \  
uses MAP (Maximum A Posteriori) probabilistic inference\footnote{Marginal inference, i.e. computing the probability of each target CE instance is also possible, but it is computationally expensive since it requires a full enumeration of a program's answer sets, or utilizing techniques for sampling from such answer sets. We are not concerned with marginal inference in this work.}, which amounts to computing a most probable answer set $\mathcal{A}$ of $\Pi = B\cup H_t \cup I_t$. From Equations (\ref{eq:weight}), 
(\ref{eq:probability}) it follows that

\begin{equation}
\label{eq:argmax}
  \mathcal{A} = \argmax\limits_{I\in \mathsf{ans}(\Pi)} P_{\Pi}(I) = \argmax\limits_{I\in \mathsf{ans}(\Pi) } W_{\Pi}(I) = \argmax\limits_{I\in \mathsf{ans}(\Pi)} \sum\limits_{r \in R_I} w_r 
\end{equation}

\noindent that is, a most probable answer set is one that maximizes the sum of weights of satisfied rules, similarly to the MLN case, for possible worlds. This is a weighted MaxSat problem that may be delegated to an answer set solver using built-in optimization tools. Since answer set solvers only optimize integer-valued objective functions, a first step is to convert the real-valued CE pattern weights to integers. We do so by scaling the weights, via multiplying them by a positive factor, while preserving their relative differences, and rounding the result to the closest integer.

Note that as it may be seen from Eq. (\ref{eq:argmax}), weight scaling by a positive factor does not alter the set of most probable answer sets. Therefore, the inference result remains unaffected, provided that rounding the weights to integer values preserves their relative differences. To ensure the latter, we set the scaling factor to $\mathit{K/d_{min}}$, where $\mathit{d_{min} = min_{i\neq j}|w_i - w_j|}$ is the smallest distance between any pair of weights and $K$ is a large positive constant, which reduces precision loss when rounding the scaled weights to integer values.

The MAP inference/weighted MaxSat computation is realized via a standard generate-and-test ASP approach, presented in Algorithm \ref{alg:map}, whose input is the background knowledge $B$, the current CE pattern set $H_t$ and the current interpretation $I_t$. First, $H_t$ is transformed into a new program, $T(H_t)$, as follows: each CE pattern $r_i$ in $H_t$ of the form $\mathit{r_i = head_i \leftarrow body_i}$ is ``decomposed'', so as to associate $\mathit{head_i}$ with a fresh predicate, $\mathsf{satisfied}/2$, wrapping $\mathit{head_i}$'s variables and its unique id, $i$ (line \ref{line:trnaform}, Algorithm \ref{alg:map}). The choice rule in line \ref{line:choice}, the ``generate'' part of the process, generates instances of $\mathsf{satisfied}/2$ that correspond to groundings of $body_i$. The weak constraint in line \ref{line:weak}, the ``test'' part of the process, decides which of the generated $\mathsf{satisfied}/2$ instances will be included in an answer set, indicating groundings of the initial CE pattern $r_i$, that will be true in the inferred state.

As it may be seen from line \ref{line:weak}, the violation of a weak constraint by an answer set $\mathcal{A}$ of $\Pi = B \cup T(H_t) \cup I_t$, i.e. the satisfaction of a ground instance of $r_i$ by $\mathcal{A}$, incurs a cost of $-w_i$ on $\mathcal{A}$, where $w_i$ is $r_i$'s integer-valued weight. The optimization process triggered by the inclusion of these weak constraints in a program generates answer sets of minimum cost. During the cost minimization process, costs of $-w_i$ are actually rewards for rules with a positive $w_i$, whose satisfaction by an answer set, via the violation of the corresponding weak constraint, reduces the answer set's total cost. The situation is reversed for rules with a negative weight, whose corresponding weak constraint is associated with a positive cost. 

 Obtaining the inferred state amounts to ``reading-off'' target CE instances from an optimal (minimum-cost) answer set of the program $B\cup T(H_t) \cup I_t$.
 
 \scriptsize
\begin{minipage}{0.46\textwidth}
	\scriptsize
	\begin{algorithm}[H]
		\caption{\scriptsize$\mathit{\mathsf{MAPInference}(B,H_t,I_t)}$ \\
			\scriptsize
			\textbf{Input: } background knowledge $B$; the current CE pattern set $H_t$; the input interpretation $I_t$. \\
			\textbf{Output: } Target CE instances included in the most probable answer set of $B\cup T(H_t) \cup I_t$.
		}\label{alg:map} 
		\scriptsize
		\begin{algorithmic}[1]
			\State $T(H_t) := \emptyset$
			\ForAll{CE pattern $r_i = \alpha \leftarrow \delta_1,\ldots,\delta_n$ in $\mathit{H_t}$ with integer weight $w_i$}
			
			\State \textbf{let} $\mathit{\mathsf{vars}(\alpha)}$ be a term wrapping the variables of $\alpha$.
			
			\State \begin{varwidth}[t]{13cm} Add to  $T(H_t)$ the following rules:\end{varwidth}
			
			\State \begin{varwidth}[t]{13cm} $\mathit{\alpha \leftarrow \mathsf{satisfied(vars}(\alpha), i)}$. \end{varwidth} \label{line:trnaform}
			
			\State $\mathit{ \{\mathsf{satisfied(vars}(\alpha), i)\} \leftarrow \delta_1,\ldots,\delta_n}.$ \label{line:choice}
			
			\State \begin{varwidth}[t]{13cm} $\mathit{ \weak \mathsf{satisfied(vars}(\alpha), i). \  [-w_i,\mathsf{vars}(\alpha), i] }$  \end{varwidth} \label{line:weak}
			
			\EndFor 
			
			\State Find an optimal answer set $\mathit{\mathcal{A}_{opt}}$ of $B\cup T(H_t) \cup I_t$.
			
			\State \textbf{return} the target CE instances in $\mathit{\mathcal{A}_{opt}}$.
		\end{algorithmic}
	\end{algorithm}
	\vspace*{4.1cm}
	\normalsize
\end{minipage}
\hfill
\begin{minipage}{0.46\textwidth}
	\begin{figure}[H]
		\centering
		\includegraphics[width=1\textwidth]{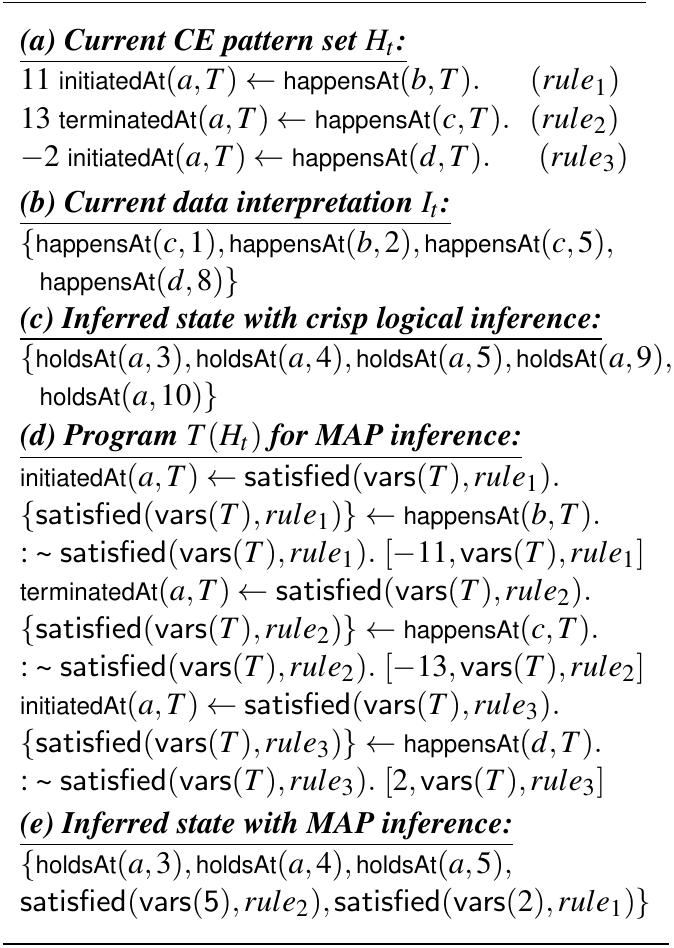}
		\vspace*{-0.6cm}
		\caption{ASP-based MAP inference example.}
		\label{fig:map}
	\end{figure} 
\end{minipage}
\normalsize

\begin{example}
\label{exmpl:map}
We illustrate the inference process via the example in Figure \ref{fig:map}, where we assume that the target CE to be recognized is $a$. 
(a) presents a CE pattern set $H_t$, i.e. a set of initiation \& termination condition for the target CE, $a$. We assume that the actual real-valued weights of the patterns have been converted into integers; (b) presents the current data interpretation $I_t$; (c) presents the inferred state obtained with crisp logical inference, i.e. the target CE instances included in the unique answer set of the program $BK\cup H_t \cup I_t$, where the CE patterns' weights have been ignored. Note that the occurrence of $\mathit{\myhappenss(b,2) \in I_t}$ initiates the target CE $a$ via $\mathit{rule_1} \in H_t$, so $a$ holds at the next time point, 3, and it also holds at time points 4 \& 5 via inertia. Then, the occurrence of $\mathit{\myhappenss(c,5) \in I_t}$ terminates $a$, via $\mathit{rule_2} \in H_t$, so $a$ does not holds at times 6,7,8, while the occurrence of $\mathit{\myhappenss(d,8) \in I_t}$ re-initiates $a$, via $\mathit{rule_3} \in H_t$, so $a$ holds at times 9 \& 10.

(d) in Figure \ref{fig:map} presents the program $T(H_t)$ obtained from $H_t$, via the transformation in Algorithm \ref{alg:map}, to allow for MAP inference;  Finally, (e) presents the MAP-inferred state, i.e. the target predicate instances included in an optimal (minimum-cost) answer set of the program $BK\cup T(H_t) \cup I_t$ (for illustrative purposes the $\mathsf{satisfied}/2$ instances in the optimal answer set are also presented). Note that the set of target CE inferences is reduced, as compared to the crisp case, since the negative-weight, $\mathit{rule_3} \in H_t$ is not satisfied by the optimal answer set. The $\mathsf{satisfied}/2$ instances in the MAP-inferred state correspond to the ground atoms $\myterminatedAts(a,5)$ and $\myinitiatedAts(a,2)$, which, along with inertia, are responsible for the target CE inferences.
\end{example}

\scriptsize
\begin{minipage}[H]{0.45\textwidth}
	\begin{algorithm}[H]
		\caption{\scriptsize $\mathit{\mathsf{LearnNewRules}(B,M,H_t,I_t,I_t^{\mathsf{MAP}}, I_t^{\mathsf{true}})}$\\
			\scriptsize
			\textbf{Input: } background knowledge $B$; mode declarations $M$, the current CE pattern set $H_t$; the current data interpretation $I_t$; the MAP-inferred state $I_t^{\mathsf{MAP}}$; the true state $I_t^{\mathsf{true}}$  \\
			\textbf{Output: } A set $\mathit{H_{new}}$ of new CE patterns.
		}
		\scriptsize
		\label{alg:new-rules} 
		\begin{algorithmic}[1]
			\State $\Pi := \emptyset$, $\mathit{H_{new}} := \emptyset$, $\mathit{H_{\bot}} := \emptyset$, $\mathit{T(H_{\bot})} := \emptyset$
			\State $\mathit{Mistakes} := I_t^{\mathsf{true}} \setminus I_t^{\mathsf{MAP}}$.
			
			\ForAll{$m \in Mistakes$}
			\State $\mathit{H_{\bot}} \leftarrow \mathsf{generateBottomRule}(m,I_t,M)$ \label{line:generate-bcs}
			\EndFor
			
			\State $\mathit{H_{\bot}} \leftarrow \mathsf{compressBottomRules}(H_{\bot})$ \label{line:compress-bcs}
			
			\ForAll{bottom rule $r_i = \alpha_i \leftarrow \delta_i^1,\ldots,\delta_i^n$ in $\mathit{H_\bot}$}\label{line:xhail-1}
			
			\State \begin{varwidth}[t]{13cm} Add to  $T(H_\bot)$ the following rules: \label{line:xhail-2}
				\par
				$\mathit{\alpha_i \leftarrow \mathsf{use}(i,0),\mathsf{try}(i,1,\mathsf{v}(\delta_i^1)),\ldots,\mathsf{try}(i,n,\mathsf{v}(\delta_i^n)).}$ 
				\par
				$\mathit{\mathsf{try}(i,1,\mathsf{v}(\delta_i^1)) \leftarrow \mathsf{use}(i,1),\delta_i^1.}$
				\par
				$\mathit{\mathsf{try}(i,1,\mathsf{v}(\delta_i^1)) \leftarrow \nbfnew \ \mathsf{use}(i,1).}$
				\par
				\ldots
				\par
				$\mathit{\mathsf{try}(i,n,\mathsf{v}(\delta_i^n)) \leftarrow \mathsf{use}(i,n),\delta_i^n.}$
				\par
				$\mathit{\mathsf{try}(i,n,\mathsf{v}(\delta_i^n)) \leftarrow \nbfnew \ \mathsf{use}(i,n).}$
			\end{varwidth}
			
			\EndFor 
			
			\State \begin{varwidth}[t]{13cm} $\Pi \leftarrow B \cup I_t \cup T(H_t) \cup T(H_\bot)$, where $T(H_t)$ is the  MAP \par inference-related transformation \par  of Algorithm \ref{alg:map} applied to the current CE pattern set $H_t$.
			\end{varwidth} \label{line:update-pi}
			
			\State \begin{varwidth}[t]{13cm} Add to $\Pi$ the following rules: 
				\par
				$\mathit{\{ \mathsf{use}(I,J) \} \leftarrow \mathsf{ruleId}(I),\mathsf{literalId}(J)}.$
				\par
				$\mathit{\weak \mathsf{use}(I,J). \ [1,I,J]}$
			\end{varwidth}\label{line:choice-weak}
			
			\State \begin{varwidth}[t]{13cm} Add to $\Pi$ one weak constraint of the form \par $\weak \mathsf{not} \  \alpha. \ [1]$
				(resp. $\weak \alpha. \ [1]$) for each target CE \par instance $\alpha$ included
				(resp. not included -- closed world \par assumption) in $I_t^{\mathsf{true}}$. 
			\end{varwidth}\label{line:weak-constr-examples}
			
			\State Find an optimal answer set $\mathit{\mathcal{A}_{opt}}$ of $\Pi$.
			
			\State Remove from $H_\bot$ every body literal $\delta_i^j$ for which $\mathsf{use}(i,j) \notin \mathcal{A}_{opt}$ and each rule $r_i$ for which 
			$\mathsf{use}(i,0) \notin \mathcal{A}_{opt}$.\label{line:synthesise-rules}
			
			\State $\mathit{H_{new}} \leftarrow H_{\bot}$.
			
			\State \textbf{return} $\mathit{H_{new}}$.
		\end{algorithmic}
	\end{algorithm}
	\normalsize
\end{minipage}
\hfill
\begin{minipage}[H]{0.5\textwidth}
	\begin{figure}[H]
		\centering
		\includegraphics[width=1\textwidth]{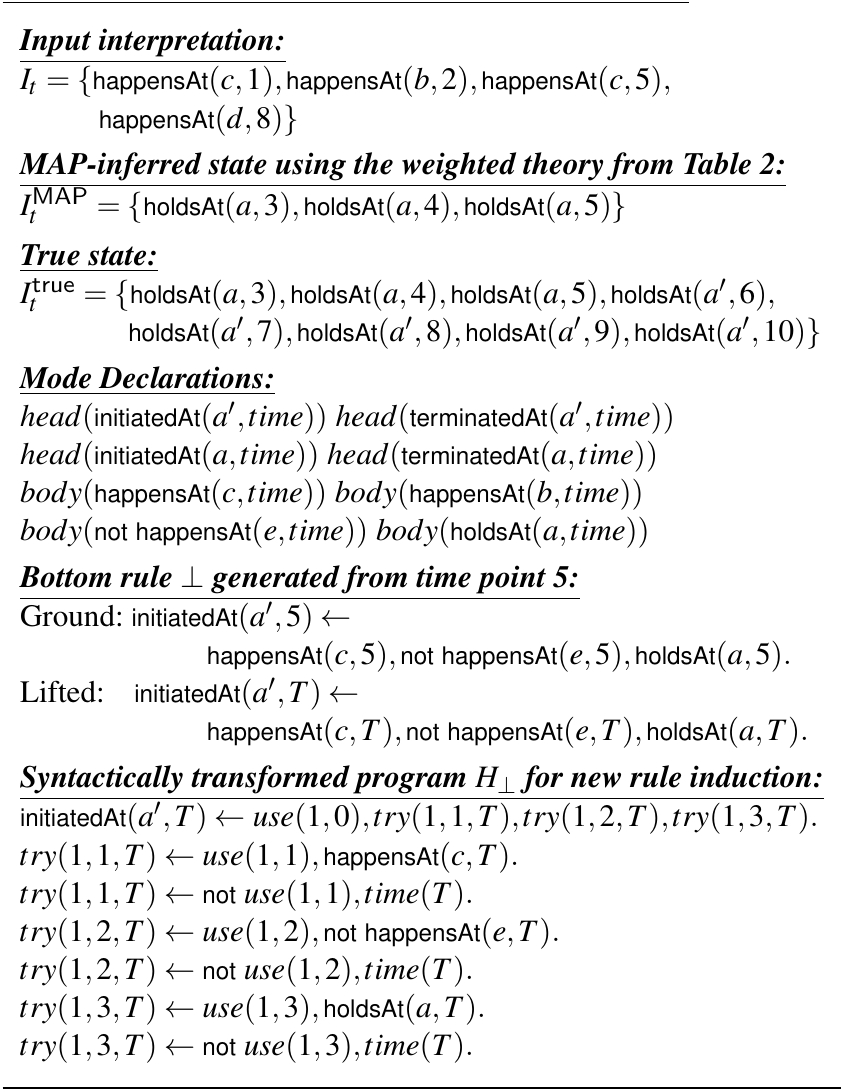}
		\vspace*{-0.6cm}
		\caption{Example of new rule induction in response to prediction mistakes.}
		\label{fig:generalization}
	\end{figure} 
	\vspace*{0.07cm} 
\end{minipage}
\normalsize

\subsection{Learning New Rules} 
\label{sec:learning-new-rules} 
Right after making a prediction via MAP inference on the current interpretation the true state is revealed to the learner, compared against the MAP-inferred state and the erroneous predictions are identified. The existing CE pattern set $H_t$ is expanded with the addition of new rules, generated in response to these mistakes, via the addition of new \myinitiatedAts/2 (resp. \myterminatedAts/2) patterns, generated from false negative (\emph{FN}) (resp. false positive (\emph{FP})) mistakes, which have the potential to prevent similar mistakes in the future. 
For instance, an \emph{FN} mistake at time $t$, i.e. a target CE instance predicted as false, while actually being true at $t$, could have been prevented via a pattern that initiates the target CE at some time prior to $t$.

Generating new CE patterns from the entirety of mistakes may result in a very large number of rules, many of which are redundant, or generated from noisy data points. To avoid that, \woled \ instantiates an optimization process that seeks a good trade-off between theory complexity and accuracy. This is done by combining structure induction techniques from non-monotonic ILP with probabilistic reasoning with the existing CE pattern set $H_t$, in order to learn a new set of CE patterns $R$, which is as compressive as possible, while at the same time, the result of reasoning with $B\cup H_t\cup R$ on the current interpetation approximates the corresponding true state as close possible. Viewing the current input interpretation as a small training set, such a process corresponds to the standard machine learning practice of jointly minimizing training error and model complexity. 

To realize this process, \woled \ employs the strategy for new rule generation, presented in Algorithm \ref{alg:new-rules}: First, a set of bottom rules (BRs) is created (line \ref{line:generate-bcs}), using the constants in the erroneously predicted atoms to generate ground \myinitiatedAts/2 and \myterminatedAts/2 atoms, which are placed in the head of a set of initially empty-bodied rules. The bodies of these rules are then populated with literals, grounded with constants that appear in the head, that are true in the current data interpretation $I_t$. The signatures of allowed body literals are specified via \emph{mode declarations} \cite{de2008logical}. 

Next, constants in the BRs are replaced by variables and the BR set is ``compressed'' (line \ref{line:compress-bcs}) to a bottom theory $H_{\bot}$, which consists of unique, w.r.t. $\theta$-subsumption, variabilized BRs. The new CE patterns are chosen among those that $\theta$-subsume $H_{\bot}$. To this end, the generalization technique of  \cite{ray2009nonmonotonic}, which allows to search into the space of theories that $\theta$-subsume $H_{\bot}$, is combined with inference with the existing weighted CE pattern set $H_t$, yielding a concise set of CE patterns $\mathit{H_{new}}$ with the property that an optimal answer set of $\mathit{B\cup H_t \cup H_{new}\cup I_t}$ best-approximates the true state associated with $I_t$.

To this end, each BR $r_i \in H_{\bot}$ is ``decomposed'' in the way shown in line \ref{line:xhail-2} of Algorithm \ref{alg:new-rules}, where the head of $r_i$ corresponds to an atom $\mathsf{use}(i,0)$ and each of its body literals, $\delta_i^j$, to a $\mathsf{try}/3$ atom, which, via the $\mathsf{try}/3$ definitions provided, may be satisfied either by satisfying $\delta_i^j$ and an additional $\mathsf{use}(i,j)$ atom, or by ``assuming'' $\mynbfs \ \mathsf{use}(i,j)$. Choosing between these two options is done via ASP optimization in line \ref{line:choice-weak} of Algorithm \ref{alg:new-rules}, where the choice rule generates $\mathsf{use}/2$ atoms that correspond to head atoms/body literals for $H_{\bot}$, and the subsequent weak constraint minimizes the generated instances to those necessary to approximate the true state, as encoded via the additional weak constraints in line \ref{line:weak-constr-examples}. New rules are ``assembled'' from the bottom rules in $H_{\bot}$, by following the prescriptions encoded in the  $\mathsf{use}/2$ atoms of an optimal answer set of the resulting program, as in line \ref{line:synthesise-rules}. 

This is essentially the \xhail \ algorithm \cite{ray2009nonmonotonic} in an ASP context. The difference of our approach from usages of this structure induction technique in previous works \cite{ray2009nonmonotonic,katzouris2015incremental}, 
is that here the search into the space of $H_{\bot}'s$ subsumers is combined with MAP inference with the existing set of weighted CE patterns (line \ref{line:update-pi}, Algorithm \ref{alg:new-rules}). Therefore, new patterns are generated only insofar they indeed help to better approximate the true state, given the already existing weighted rules. This technique allows to generalize from the data in the current interpretation, while taking into account previously discovered patterns and their relative quality, as reflected by their weights.  

\begin{example}
	\label{ex:exmpl2}
We illustrate the rule induction technique in Algorithm \ref{alg:new-rules} via an example. Recall Example \ref{exmpl:map} and assume that after making a prediction and generating the MAP-inferred state we receive the true state presented in Figure \ref{fig:generalization}. Assume also that we have an additional target CE here, $a'$. The true state in Figure \ref{fig:generalization} differs from the MAP-inferred one as it contains \myholdsAts/2 atoms concerning $a'$. These atoms are missing from the MAP-inferred state, therefore, they are false negative (\emph{FN}) predictions, i.e. true instances of a target complex event, which are not recognized by the weighted theory of Figure \ref{fig:map}. In order to revise the current theory towards eliminating the \emph{FN}'s we employ the strategy of Algorithm \ref{alg:new-rules}. First, a set of ground atoms that eliminate the erroneous predictions if added to the current theory are generated via abductive reasoning. These atoms will serve as heads for new rules and their signatures are predefined via mode declarations (see Figure \ref{fig:generalization}). In our example one such atom, $\myinitiatedAts(a',5)$, suffices (note that $a'$ in the true state holds from time 6 onwards, therefore, it must have been initiated in the previous time point). A bottom rule $\bot$ having this atom in the head is generated from the data in the current interpretation, as shown in Figure \ref{fig:generalization}. The signatures of literals in the body of $\bot$ are specified via the mode declarations and the actual atoms are groundings of such literals, generated using constants in the head, that are true in the data. The ``lifted'' version of $\bot$ shown in Figure \ref{fig:generalization} is then transformed into  program $H_\bot$ in Figure \ref{fig:generalization}, forming a search space for learning a new rule.   

Approximating the true state is then realized via finding an optimal answer set of a program $\Pi$ consisting of the following parts: (i) the axioms of the Event Calculus (background knowledge); (ii) program $T(H_t)$ from Figure \ref{fig:map}, used for probabilistic inference with the weighted theory $H_t$; (iii) program $H_\bot$ from Figure \ref{fig:generalization}, used for new rule induction; (iv) a choice rule for $\mathit{use/2}$ atoms and weak constraints that minimize the $\mathit{use/2}$ atoms included in the optimal answer set, thus encoding a preference for simpler rules in the rule induction process (line \ref{line:choice-weak}, Algorithm \ref{alg:new-rules}); (v) weak constraints that minimize the ``disagreement'' between the optimal answer set and the true state w.r.t. the target CE instances included in the optimal answer set (line \ref{line:weak-constr-examples}, Algorithm \ref{alg:new-rules}).

An optimal answer set of program $\Pi$ contains the atoms $\mathit{use(1,0), use(1,1)}$ and $\mathit{use(1,3)}$, which correspond to the rule $r^{\star}: \mathit{\myinitiatedAts(a',T)\leftarrow \myhappenss(c,T),\myholdsAts(a,T)}$. This is the simplest rule $r$ with the property that $H_t \cup r$ correctly accounts for the true state. 

\noindent \textbf{Remark.} Note that reasoning with the weighted theory $H_t$ is necessary in order to learn the new rule $r^{\star}$ in this example. Indeed, $\mathit{rule_1}$ in Figure \ref{fig:map} initiates $a$ at time 2, thus causing it to hold at time 5 and allowing the $\mathit{\myholdsAts(a,T)}$ atom to be added to the body of $r^{\star}$. Observe that without this atom, the more general rule, $\mathit{\myinitiatedAts(a',T)\leftarrow \myhappenss(c,T)}$ would yield a number of false positive predictions, since it would initiate $a'$ at time 1, due to the $\mathit{\myhappenss(c,1)}$ atom in the input data, thus causing $a'$ to (erroneously) hold in the interval $[2,5]$. It can be seen by comparing costs in the optimization process that ``settling'' for the initial false negative predictions for $a'$ in the interval $[6,10]$ (by not learning a new rule at all) is more cost-efficient than learning the rule $\mathit{\myinitiatedAts(a',T)\leftarrow \myhappenss(c,T)}$, which does retrieve the false negatives, but causes the false positives for $a'$ in the interval $[2,5]$. Therefore, had we not used $H_t$ as background knowledge, we would not have learned a new rule from the data in this example. It is worth mentioning that the same would have happened if we used $H_t$ as a crisp theory, i.e. without the rules' weights. In that case $\mathit{rule_3}$ from Figure \ref{fig:map} would be responsible for a number of false positives in the inferred state. This example, therefore, demonstrates the merit of combining probabilistic inference with existing rules with the new rule induction process. 
\end{example}

\subsection{Weight Learning}
\label{sec:weight-learning}
The CE patterns' weights (which are initialized to a close-to-zero value) are updated by comparing their true groundings in the inferred and the true state respectively. For a target $CE$ $\alpha$ and an $\myinitiatedAts/2$ (resp. $\myterminatedAts/2$) CE pattern $r_i$, a true grounding, either in the inferred, or in the true state, is a grounding of $r_i$ at time $t$, such that $\myholdsAts(\alpha,t+1)$ is true (resp. false). CE patterns that contribute towards correct predictions are promoted, while those that contribute to erroneous predictions are down-weighted. 

As in \cite{DBLP:conf/pkdd/KatzourisMAP18}, we use the \adagrad \ algorithm \cite{duchi2011adaptive} for weight updates, a version of Gradient Descent that dynamically adapts the learning rate, i.e. the magnitude of weight promotion/demotion, for each CE pattern individually, by taking into account the pattern's performance on the past data. \adagrad \ updates a weight vector, whose coordinates correspond to a set of features (the CE patterns in our case), based on the subgradient of a convex loss function of these features. Our loss function, called the \emph{prediction-based loss}, is a simple variant of the max-margin loss for structured prediction for MLNs \cite{huynh2011online}, whose subgradient is the vector with $\Delta g_i$ in its $i$-th coordinate, where $\Delta g_i = g_{r_i}^{MAP} -  g_{r_i}^{true}$ denotes the difference in the true groundings of the $i$-th pattern, $r_i$, in the MAP-inferred and the true  state respectively. Essentially, $\Delta g_i$ counts prediction mistakes which are relevant to $r_i$, either false positive mistakes ($\Delta g_i > 0$), for which  $r_i$ is responsible, or false negative predictions ($\Delta g_i < 0$), committed by the entire theory, which $r_i$ could have helped prevent. The weight update rule for the $i$-th CE pattern $r_i$ is then:     

\small
\begin{equation}
\label{eq:adagrad}
w_i^{t+1} = sign(w_i^{t} - \frac{\eta}{C_i^t}\Delta g_i^t)  \  max\{0, |w_i^t - \frac{\eta}{C_t^i}\Delta g_i^t| - \lambda \frac{\eta}{C_t^i} \}
\end{equation}
\normalsize

\noindent where $\eta$ is a learning rate parameter, $\lambda$ is a regularization parameter and $C_i^t = \delta + \sqrt{\sum_{j=1}^{t}(\Delta g_i^j)^2}$ is a term proportional to the sum of $r_i$'s past subgradients (the $\Delta g_i$'s) (plus a $\delta \geq 0$ to avoid division by zero in $\eta/C_i^t$). Note that according to Eq. (\ref{eq:adagrad}), a $\Delta g_i > 0$, i.e. a case where $r_i$ is responsible for false positive predictions, leads to a weight demotion for $r_i$, while a $\Delta g_i < 0$ leads to a promotion of its weight, that can help the theory retrieve the false negative misses. 

The $C_i^t$ term is the adaptive factor that assigns a different learning rate to each CE pattern, since the magnitude of a weight update via the term $|w_i^t - \frac{\eta}{C_t^i}\Delta g_i^t|$ is affected by the CE pattern's previous history, in addition to its current utility, expressed by $\Delta g_i^t$. Smaller values for $C_i^t$ correspond to ``rare'', but potentially highly informative features, and therefore lead to weight updates of larger magnitude. The ``informativeness'' of these features is reflected in the magnitude of the current subgradient $\Delta g_i$, since, regardless of the value of $C_i^t$, zero, or very small values of the current subgradient (corresponding to non-informative features) have very small effect to the weight.   

The regularization term in Eq. (\ref{eq:adagrad}), $\lambda \frac{\eta}{C_t^i}$, is the amount by which the $i$-th CE pattern's weight is discounted when $\Delta g_i^t = 0$. As usual, the role of regularization is to introduce a bias towards simpler models, in this case by eventually (over time) pushing to zero the weights of irrelevant rules, that play no significant role in helping the theory make correct predictions.

 \subsection{Revising Existing CE patterns' Structure}
\label{sec:specialization}
Although the rules induced with the process described in Section \ref{sec:learning-new-rules} are useful locally, that is, w.r.t. the current input interpretation $I_t$, they may be proven inadequate  w.r.t. a more ``global'' view of the data. A case where this typically occurs is inducing the simplest rule that helps approximate the true state in $I_t$, which turns out to be over-general once larger regions of the data are taken into account. Weight learning may help suppress the effects of such rules in the predictive performance of the model, but this is not enough: Since the data are processed incrementally in small batches, it may be the case that a ``data view'' large-enough for learning a high-quality target rule may never occur, causing the learning process to fail.    

A remedy is to revise the rules' structure over time, as larger portions of the data are revealed. Similarly to \oled \ \cite{DBLP:journals/tplp/KatzourisAP16}, \woled \ does so via  
a classical in ILP, hill-climbing search process, searching for a high-quality CE pattern into a \emph{subsumption lattice} defined by a bottom rule. Such bottom rules are generated during the rule induction process of Section \ref{sec:learning-new-rules}. Each induced rule $r$ is associated with a bottom rule $\bot_r$ (such that $r$ $\theta$-subsumes $\bot_r$), which serves as a pool to draw literals from, in order to specialize $r$ over time. This process is online, using the data in the incoming interpretations to evaluate a CE pattern and its current specializations. A Hoeffding test \cite{DBLP:conf/kdd/DomingosH00} allows to identify, with high probability, the best specialization from a small subset of the input interpretations. Once the test succeeds, the parent rule is replaced by its best specialization and the process continues for as long as new specializations improve the current rule's performance.
 
In particular, at each point in time a parent rule and its specializations are evaluated on incoming data, via an information gain scoring function, assessing the cumulative merit of a specialization over the parent rule, across the portion of the stream seen so far:

\vspace*{-0.5cm}

\begin{equation*}
G(r,r') = P_r \cdot (log\frac{P_r}{P_r+N_r} - log\frac{P_{r'}}{P_{r'}+N_{r'}})
\label{eq:foil-gain}
\end{equation*}

\noindent where $r'$ is $r$'s parent rule and for each rule $r$, $P_r$ (resp. $N_r$) denotes the sum of true (resp. false) groundings of $r$ in the MAP-inferred states generated so far. The information gain function is normalized in $[0,1]$ by taking $0$ as the minimum (as we are interested in positive gain only) and dividing a $G$-value by its maximum, $G_{max}(r,r') = P_{r'} \cdot (-log \frac{P_{r'}}{P_{r'}+N_{r'}})$. When the range of $G$ is $[0,1]$, a Hoeffding test succeeds, allowing to select $r_1$ as the best of a parent rule $r$'s specializations, when $G(r_1,r) - G(r_2,r) > \epsilon = \sqrt{\frac{log 1/\delta}{2N}}$, where $r_1, r_2$ are respectively $r$'s best and second-best specializations, $\delta$ is a confidence parameter and $N$ is the number of observations seen so far, we refer to \cite{DBLP:journals/tplp/KatzourisAP16} for further details. 

\begin{figure}[t]%
	\centering
	\subfloat{{\includegraphics[width=6.5cm]{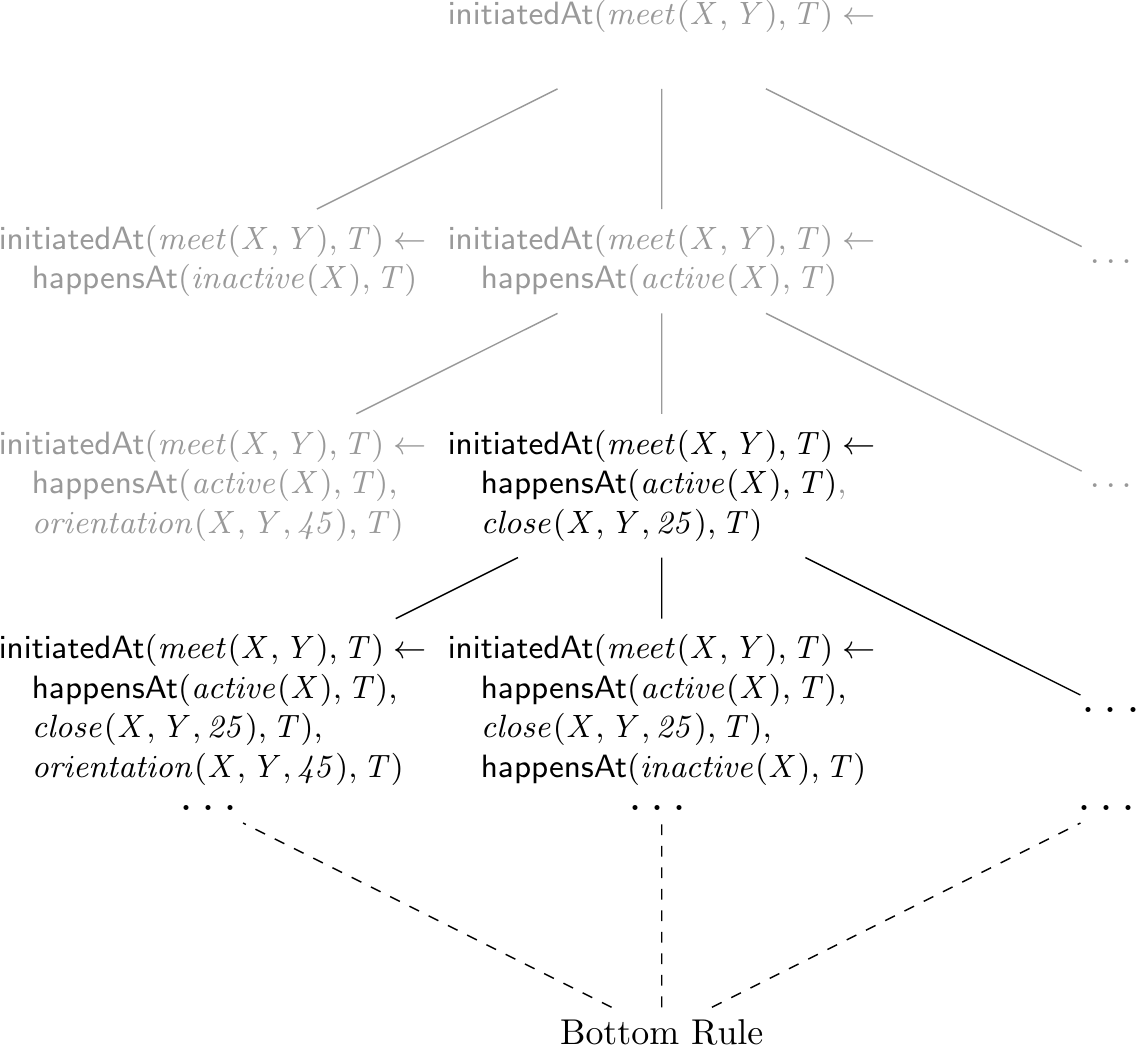} }}
	\caption{\small A subsumtion lattice.}%
	\label{fig:subsumption-lattice}%
\end{figure}

A successful Hoeffding test results in replacing the parent rule $r$ with its best specialization $r_1$ and moving one level down in the subsumption lattice, via generating $r_1$'s specializations and subsequently evaluating them on new data. Figure \ref{fig:subsumption-lattice} illustrates the process for an initiation CE pattern. The rules at each level of the lattice represent the specializations of a corresponding rule at the preceding level. The greyed-out part of the search space in Figure \ref{fig:subsumption-lattice} represents the portion that has already been searched, while the non greyed-out rule at the third level represents the best-so-far rule that has resulted from a sequence of Hoeffding tests. 

The specializations' weights are learnt simultaneously to those of their parent rules, by comparing the specializations' true groundings over time in the MAP-inferred states (generated from ``top theories'', consisting of parent rules only) and the true states respectively.    

\section{Discussion on Implementations}
\label{sec:implementations}
We highlight the differences between the ASP-based version of \woled, which we henceforth denote by \woledasp, with the version of \cite{DBLP:conf/pkdd/KatzourisMAP18}, which relies on MLN libraries, and which we henceforth denote by \woledmln.

Contrary to \woledasp, which is based entirely on the \clingo\footnote{\url{https://potassco.org/}} answer set solver, \woledmln \ is based on a number of different software tools. It uses the \lomrf \ library for Markov Logic Networks \cite{LoMRF}, for grounding MLN theories and performing circumscription via predicate completion \cite{skarlatidis2015probabilistic}, in order to convert them into a form that supports the non-monotonic semantics of the Event Calculus for reasoning, something that \woledasp \ has out of the box. MAP inference in \woledmln \ is performed via a state-of-the-art in MLNs, Integer Linear Programming-based approach, which is introduced in \cite{huynh2009max} and is implemented using the \lpsolve\footnote{\url{https://sourceforge.net/projects/lpsolve}} \ solver.

Another important difference between \woledasp \ and \woledmln \ from an algorithmic perspective, lies in the new CE pattern generation process. As discussed in Section \ref{sec:learning-new-rules}, \woledasp \ is able to perform the search for new structure, while taking into account the contribution of the weights of existing patterns in approximating the true state. In contrast, \woledmln \ lacks this ability. It generates a bottom theory $H_{\bot}$ from the erroneously predicted atoms, and then attempts to gradually learn a high-quality CE pattern from the rules therein, regardless of their quality. In comparison, \woledasp's strategy may lead, in principle, to simpler theories of more meaningful rules.

\section{Experimental Evaluation}
\label{sec:experiments}

We present an experimental evaluation of our approach on three CER data sets from the domains of \emph{activity recognition, maritime monitoring} and \emph{vehicle fleet management}. 

\noindent \textbf{Datasets.} \emph{CAVIAR}\footnote{\url{http://homepages.inf.ed.ac.uk/rbf/CAVIARDATA1/}} is a benchmark dataset for activity recognition, described in Section \ref{sec:background}, consisting of 28 videos with 26,419 video frames in total. We experimented with learning CE patterns for two CEs from CAVIAR, related to two people \emph{meeting each other} and \emph{moving together}, which we henceforth denote by \emph{meeting} and \emph{moving} respectively. There are 6,272 video frames in CAVIAR where \emph{moving} occurs and 3,722 frames where \emph{meeting} occurs. 
A fragment of a CE definition for \emph{moving} is presented in Table \ref{table:stream1}(d).

Our second dataset is a publicly available dataset from the field of maritime monitoring\footnote{\url{https://zenodo.org/record/1167595\#.WzOOGJ99LJ9}}. It consists  of  Automatic  Identification  System (AIS) position signals collected from vessels sailing around the area of Brest, France, for a period of six months, between October and March 2015. The data have been pre-processed using trajectory compression techniques \cite{patroumpas2017online} that identify ``critical points'' in a trajectory, i.e. mobility features, such as vessel stops, turns, slow motion movements etc. The critical points maritime dataset has been further pre-processed, in order to extract spatial relations between vessels (e.g. vessels being close to each other) and areas of interest, such as protected areas, areas near coast, open-sea areas etc. There are 16,152,631 critical points in the maritime dataset, involving 4,961 vessels and 6,894 areas, for a total size of approximately 1,3GB. 

The maritime dataset is not labeled in terms of occurring CE instances, we therefore used hand-crafted CE patterns to perform CER on the critical points, thus generating the annotation. The purpose of learning was to reconstruct the hand-crafted CE patterns. We experimented with learning CE patterns for a CE related to vessels involved in potentially suspicious rendezvous (henceforth denoted by \emph{rendezVous}), which holds when two vessels are stopped, or move with very low speed in proximity to each other in the open sea. Since such behavior is often related to illegal activities, tracking it is of special interest for maritime surveillance. 

Our third dataset is provided by Vodafone Innovus\footnote{\url{https://www.vodafoneinnovus.com}}, a commercial vehicle fleet management provider and our partner in the Track \& Know\footnote{\url{https://trackandknowproject.eu/}} EU-funded funded project. The data consist of time-stamped vehicle positions (GPS), in addition to mobility-related events, such as \emph{abrupt acceleration, abrupt deceleration, harsh cornering}, provided  by an accelerometer device installed in each commercial vehicle. Moreover, map-matched weather attributes were used to enrich the dataset with contextual information, such as \emph{icy road}. We refer to \cite{DBLP:journals/tplp/TsilionisKNDA19} for a detailed account of this dataset.

Similarly to the maritime dataset, due to the lack of CE-related ground truth we used hand-crafted patterns, developed in collaboration with domain experts in Track \& Know to generate the ground truth CE instances.
The learning target was a CE related to \emph{dangerous driving}, which holds in a number of occasions, such as abruptly accelerating/decelerating on an icy road. The fleet management dataset consists of 4M records for a total size of 527 MB.     

All experiments were carried-out on a machine with a 3.6GHz processor (4 cores, 8 threads) and 16GB of RAM. \clingo \ (v. 5.4.0) was used with the \textsf{\footnotesize --opt-strategy=usc} option, which significantly speeds-up the optimization process. The hyper-parameters for the different algorithms the are compared in these experiments were set as follows: For \adagrad, $\eta = 1.0, \lambda = 0.01, \delta = 1.0$. The significance parameter for Hoeffding tests was set to $\delta = 10^{-2}$. The code for all algorithms used in these experiments in available online\footnote{\url{https://github.com/nkatzz/ORL}}.

\subsection{Scalability of Inference}

The purpose of our first experiment was to compare the scalability of the ASP-based MAP inference process, which lies at \woledasp's core, to that of \woledmln's.  To that end we used the task of online weight learning with hand-crafted CE patterns, where the learner is required to first perform MAP inference on the incoming interpretations with a fixed-structure CE pattern set, and then update the CE patterns' weights, based on their contribution to erroneous inferences in the MAP-inferred state. 
Given that the weight update cost is negligible and the CE pattern set is fixed, the MAP inference cost is the dominant one in this task and it depends on the cost of grounding the current CE pattern set, plus the cost of solving the corresponding weighted MaxSat problem for each incoming interpretation. Note that since the CE pattern sets for each CE are fixed in this experiment, predicate completion in \woledmln \ is performed only once at the beginning of a run, therefore its cost is negligible.

\begin{figure*}[t!]
	\renewcommand{\thesubfigure}{a}
	\subfloat[(a) \emph{Meeting} ]
	{\centering
		\includegraphics[width=0.25\textwidth]{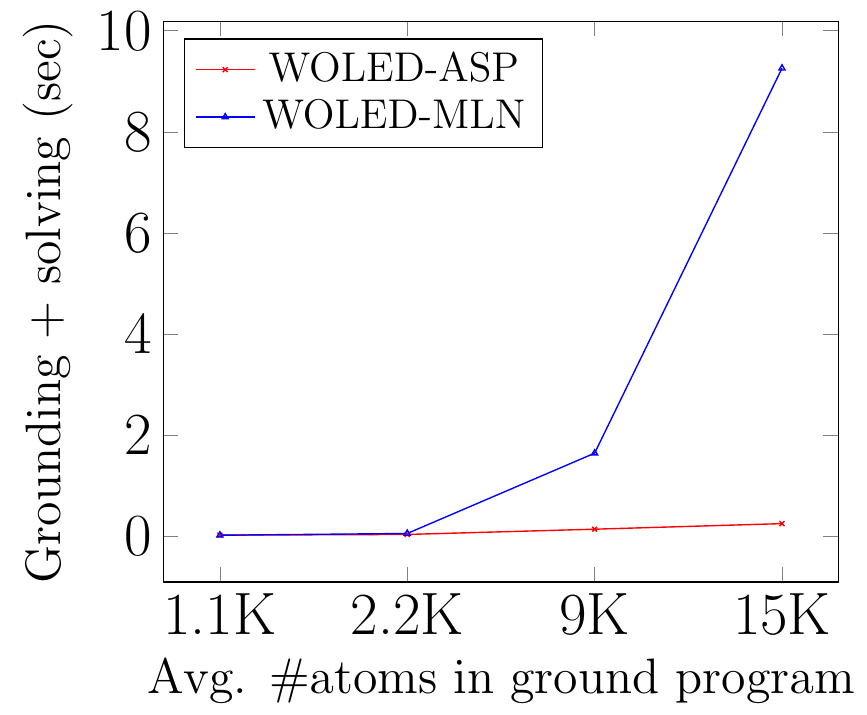}
		\label{fig:meeting}}
	\hspace*{-0.3cm}
	\renewcommand{\thesubfigure}{b}
	\subfloat[(a) \emph{Moving} ]
	{\centering
		\includegraphics[width=0.25\textwidth]{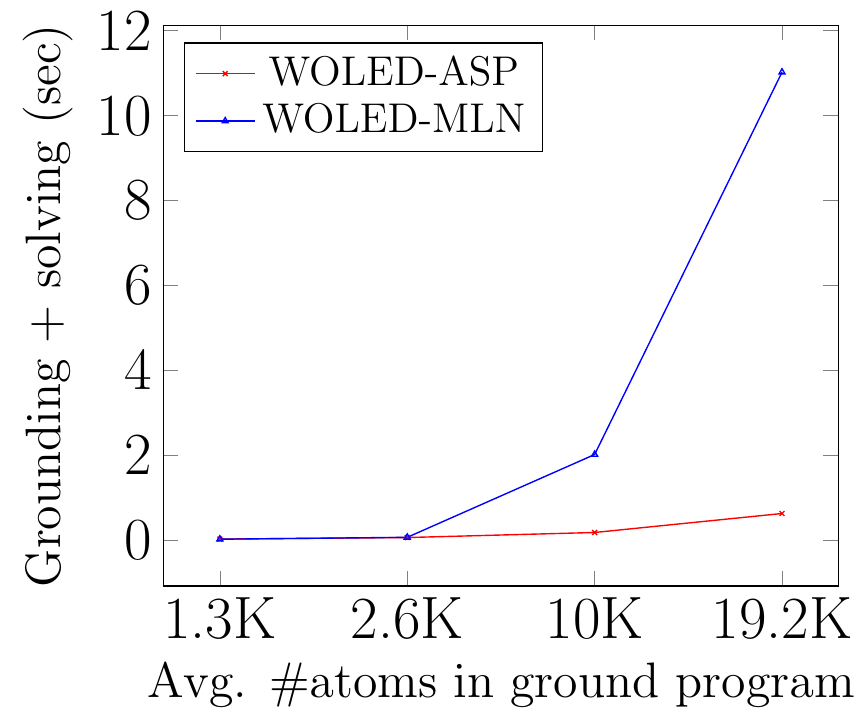}
		\label{fig:moving}}
	\hspace*{-0.3cm}
	\renewcommand{\thesubfigure}{c}
	\subfloat[(c) \emph{Rendezvous} ]
	{\centering
		\includegraphics[width=0.25\textwidth]{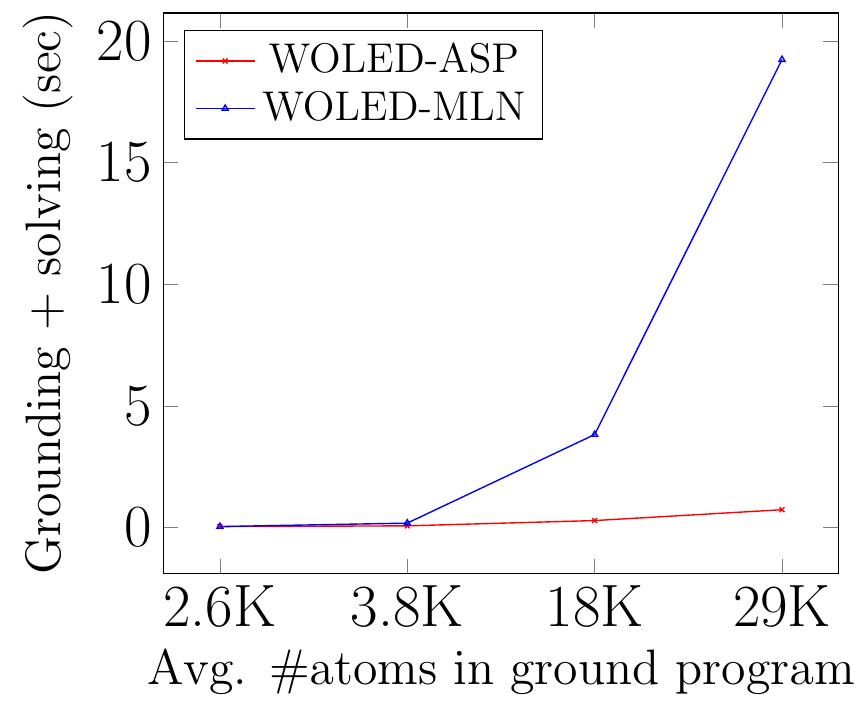}
		\label{fig:rendezvous}}
	\hspace*{-0.3cm}
	\renewcommand{\thesubfigure}{d}
	\subfloat[(d) \emph{Dangerous driving} ]
	{\centering
		\includegraphics[width=0.25\textwidth]{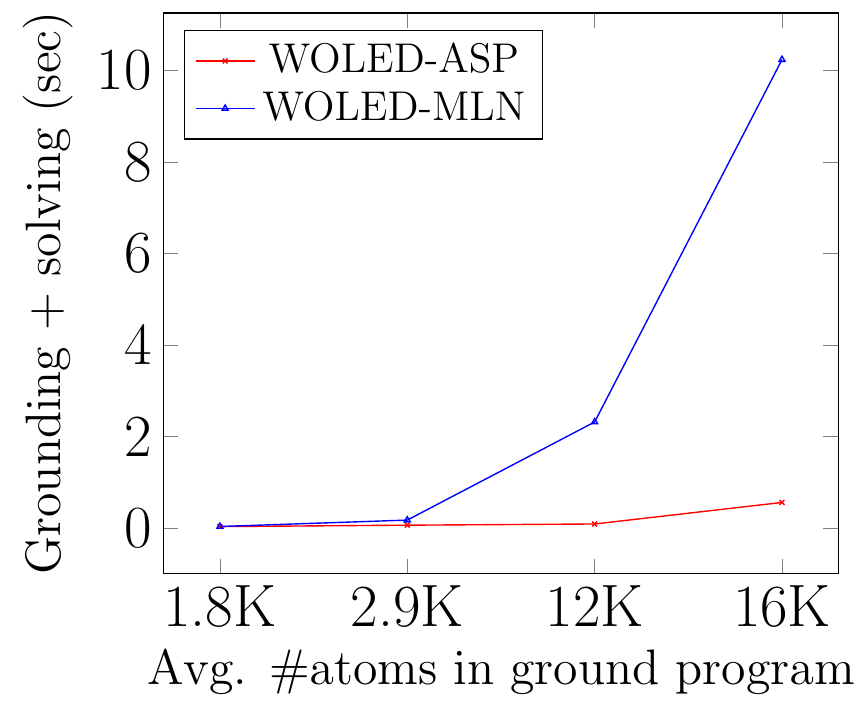}
		\label{fig:dang-driving}}
	\caption{\small Scalability of MAP inference.
	}
	\label{fig:map-scalability}
\end{figure*} 

The data were consumed by the learners in mini-batches, where each mini-batch is an interpretation consisting of data in a particular time interval. We performed weight learning with different mini-batch sizes of 50, 100, 500 \& 1000 time points. We measured the average MAP inference time (grounding plus solving time) for \woledasp \ and \woledmln \ respectively, throughout a single-pass over the data, for different mini-batch sizes. Note that as the mini-batch size grows, so does the size of the corresponding ground program from which the MAP-inferred state is extracted.    

Figure \ref{fig:map-scalability} presents the results, which indicate that the growth in the size of the ground program, as the mini-batch size increases, entails an exponential growth to the MAP inference cost for \woledmln. In contrast, thanks to \clingo's highly optimized  grounding and solving abilities, MAP inference with \woledasp \ takes near-constant time.

\subsection{Online Structure \& Weight Learning Performance}
\label{sec:online-performance}

In our next experiment we assess \woledasp's predictive performance and efficiency in the task    
of online structure \& weight learning and we compare it to (i) \woledmln; (ii) \oled \ \cite{DBLP:journals/tplp/KatzourisAP16}, the crisp version of the algorithm that learns unweighted CE patterns; (iii) \hand, a set of predefined rules for each CE and (iv) \hand-WL, the rules in \hand \ with weights learnt by \woledasp.

To assess the predictive performance of the systems compared we used two methods: \emph{Prequential} evaluation \cite{bifet2018machine}, where each incoming data interpretation is first used to evaluate the current CE pattern set and then to update its structure and weights, and standard cross-validation. In prequential evaluation we typically measure the average prediction loss over time, which is an indication of a learner's ability to incorporate new information that arrives over time into the current model. With cross-validation we assess a learner's generalization abilities, by evaluating the predictive performance of a learnt model on a test set.

\footnotesize
\setlength{\tabcolsep}{3pt}
\begin{table*}[t]
	\scriptsize
	\begin{minipage}{\textwidth}
		\begin{tabular}{>{\centering\arraybackslash}p{1.5cm}>{\centering\arraybackslash}p{1.5cm}>{\centering\arraybackslash}p{2cm}>{\centering\arraybackslash}p{2cm}>{\centering\arraybackslash}p{1.5cm}>{\centering\arraybackslash}p{2cm}>{\centering\arraybackslash}p{1.5cm}}
			\hline

			~ & \scriptsize\textbf{Method} & $\mathbf{F_1}$\textbf{-score (test set)} & \textbf{Theory size} & \textbf{Inference Time (sec)} & \textbf{Pred. Compl. Time (sec)} & \textbf{Total Time (sec)}\\ 
			\noalign{\smallskip}
			
			\emph{Moving} & \tiny \textsf{WOLED-ASP} & \textbf{0.821} & 26 & 15 & -- & 112 \\
			
			~& \tiny \textsf{WOLED-MLN} & 0.801 & 47 & 187 & 28 & 478 \\
			
			~& \tiny \textsf{OLED} & 0.730 & \textbf{24} & \textbf{13} & -- & 74 \\
			
			~& \tiny \textsf{HandCrafted} & 0.637 & 28 & -- & -- & -- \\
			
			~& \tiny \textsf{HandCrafted-WL}  & 0.702 & 28 & 16 & -- & \textbf{52} \\
			
			\hline
			
			\emph{Meeting} & \tiny \textsf{WOLED-ASP} & \textbf{0.887} & 34 & \textbf{12} & -- & 82 \\
			
			~& \tiny \textsf{WOLED-MLN} & 0.841 & 56 & 134 & 12 & 145 \\
			
			~& \tiny \textsf{OLED} & 0.782 & 42 & 10 & -- & 36 \\
			
			~& \tiny \textsf{HandCrafted} & 0.735 & \textbf{23} & -- & -- & -- \\
			
			~& \tiny \textsf{HandCrafted-WL} & 0.753 & \textbf{23} & 13 & -- & \textbf{31} \\
			
			\hline
			
			\emph{Rendezvous} & \tiny \textsf{WOLED-ASP} & 0.98 & 18 & 647 & -- & 4,856 \\
			
			~ & \tiny \textsf{WOLED-MLN} & 0.98 & 18 & 2,923 & 434 & 6,218 \\
			
			~ & \tiny \textsf{OLED} & 0.98 & 18 & \textbf{623} & -- & \textbf{4,688} \\
			
			
			
			\hline
			
			\emph{Dang.Drive} & \tiny \textsf{WOLED-ASP} & 0.99 & \textbf{21} & 341 & -- & 2,465 \\
			
			~& \tiny \textsf{WOLED-MLN} & 0.99 & 28 & 926 & 287 & 3,882 \\
			
			~& \tiny \textsf{OLED} & 0.99 & \textbf{21} & \textbf{312} & -- & \textbf{2,435} \\
			
			\hline
		\end{tabular}
	\end{minipage}
	\caption{Online structure \& weight learning results.}\label{table:results}
\end{table*}
\normalsize

\begin{figure*}[t!]
	\renewcommand{\thesubfigure}{a}
	{\centering
		\includegraphics[width=.36\textwidth]{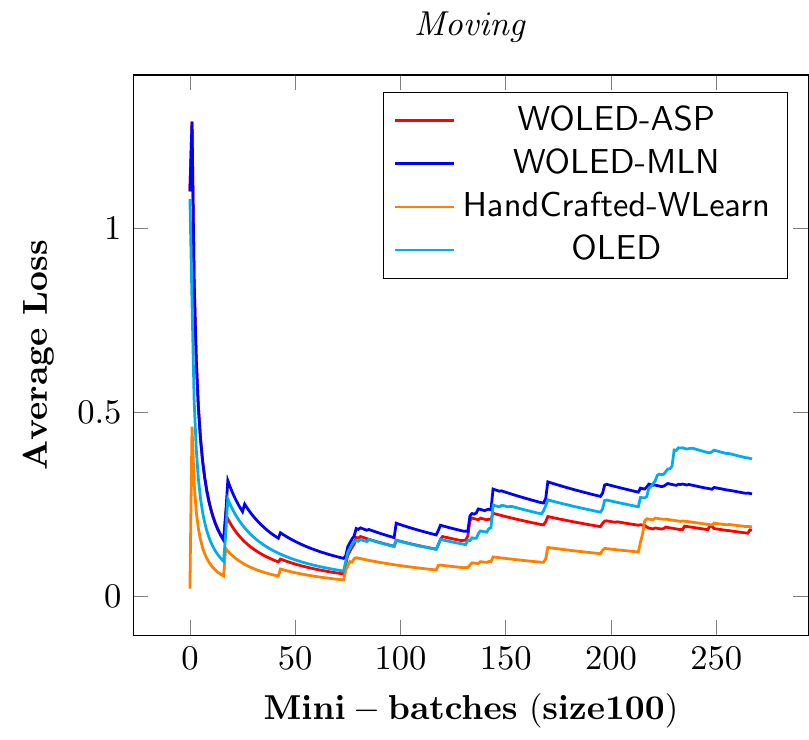}
	}
	\renewcommand{\thesubfigure}{b}
	{\centering
		\includegraphics[width=.36\textwidth]{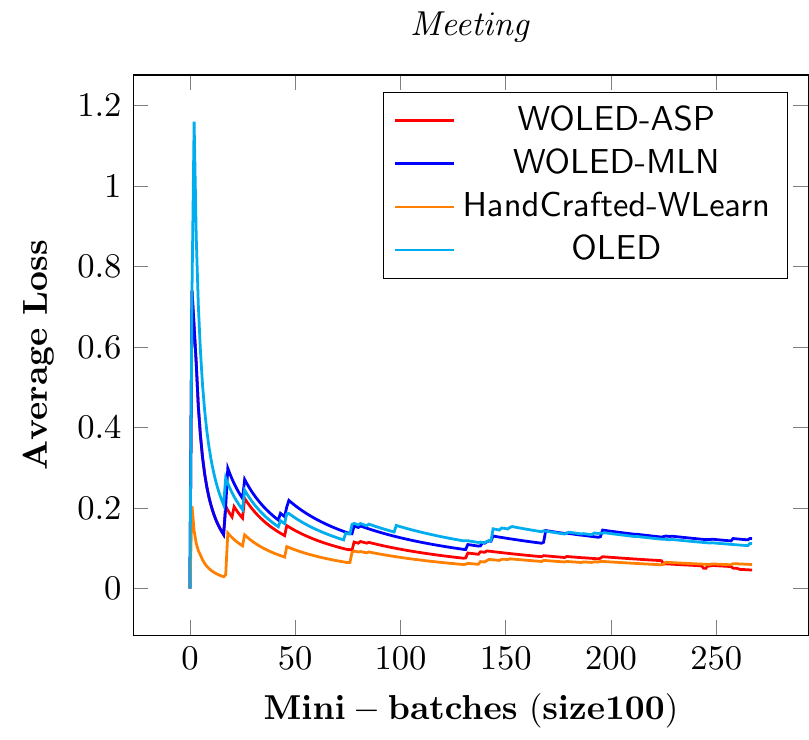}
	}
	\caption{\small Prequential Evaluation on CAVIAR.
	}
	\label{fig:prequential}
\end{figure*} 

The results from prequential evaluation for \emph{meeting} \& \emph{moving} are presented in Figure \ref{fig:prequential}, while Table \ref{table:results} reports several statistics for the systems being compared: (i) $F_1$-scores on a test set. For CAVIAR we used tenfold cross-validation and the reported $F_1$-scores are micro-averages obtained from ten different test sets. For the maritime and the fleet management datasets, whose size makes tenfold cross-validation impractical, we used half the dataset for training and half for testing, so the reported $F_1$-scores are obtained from the latter half; (ii) CE pattern set sizes (total number of literals) at the end of a single-pass over a dataset; (iii) total inference time at the end of a single-pass over a dataset (MAP inference for \woledasp, \woledmln \ \& \hand-WL, crisp logical inference for \oled); (iv) for \woledmln, total time spent on predicate completion;  
(v) total training time at the end of a single-pass over a dataset, which includes time spent on CE pattern generation, computing $\theta$-subsumption etc, i.e. the dominant costs involved in learning CE patterns structure. Note that we report on (iii), (iv), (v) only for approaches that require training (i.e., not for \hand). Also, we did not experiment with hand-crafted CE patterns in the maritime and the fleet management datasets, since in these datasets hand-crafted CE patterns were used to generate the ground truth in the first place. We also omit prequential learning curves for \emph{rendezVous} \& \emph{dangerous driving} in Figure \ref{fig:prequential}, since, due to the synthetic ground truth in the maritime \& the fleet management datasets, the learning curves for these complex events are very similar for all algorithms being compared and are not informative.   

The results in Figure \ref{fig:prequential} and in Table \ref{table:results} seem to validate the claim of Section \ref{sec:implementations} on the differences in predictive performance between \woledasp \ and \woledmln. Indeed, \woledasp \ clearly outperforms \woledmln, both in prequential error and in cross-validation $F_1$-scores, indicating better generalization abilities. Moreover, \woledasp \ learns simpler CE patterns sets, as shown by the theory size statistic. \hand-WL outperforms \woledasp \ in the prequential task for the most part of the training process. This was expected, since \ \hand-WL has the advantage of operating on a good set of rules provided beforehand and, therefore, is less prone to erroneous predictions. On the other hand, its inability to learn new rules explains its inferior test-set $F_1$-scores for \emph{meeting} \& \emph{moving}. 

\oled \ aims at quickly discovering a good set of rules. It is not concerned with optimizing their joint predictive performance and does not learn weights. It is the most efficient of all learners, but it is also outperformed by all in terms of prequential error and test set $F_1$-scores.

Regarding efficiency, it may be seen by comparing inference times to total training times, that the dominant cost is related to structure learning tasks (recall that total training times factor-in such costs). Yet, in comparison to \woledmln, \woledasp \ achieves significantly lower costs for MAP inference, which approximate the cost of \oled's crisp logical inference. In addition to its more sophisticated CE pattern creation strategy, which tends to generate fewer CE patterns of high quality, this results in \woledasp \ being significantly more efficient than \woledmln. Note also, that an additional, not negligible cost for \woledmln \ stems from predicate completion.   

\subsection{Comparison to Batch Learners}
\label{sec:batch-performance}

In our last experiment we compare \woledasp \ to a number of batch learning algorithms for learning structure and weights. To that end we used smaller datasets whose size makes batch learning practical. In particular, we used a fragment of the CAVIAR dataset that has been used in batch learning experiments in previous work \cite{skarlatidis2015probabilistic} and a small excerpt of the maritime dataset. The fragment CAVIAR dataset consists of the regions of the original dataset where the two target complex events (\emph{meeting} and \emph{moving}) occur and contains a total of 25,738 training interpretations. The maritime fragment dataset contains 11,930 training interpretations corresponding to six data sequences, extracted from the original maritime dataset, where \emph{rendezvous} between pairs of vessels occurs. 

We compare \woledasp \ to the following batch learning algorithms: (i) \xhail \ \cite{ray2009nonmonotonic}, a non-monotonic ILP learner whose rule induction strategy \woledasp \ combines with probabilistic reasoning; (ii) \iled \ \cite{katzouris2015incremental}, an algorithm that combines \xhail's learning machinery with theory revision, in order to learn incrementally, and has been shown to achieve performance comparable to that of \xhail's, while being much more efficient; (iii) \ilasp \  \cite{ILASP_system}, a state of the art ILP system that learns answer set programs from examples and has been used for inducing complex event patterns \cite{ilasp}; (iv) \maxmargin, a batch weight learning algorithm for MLN, introduced in \cite{huynh2009max}, which has been used with the CAVIAR fragment dataset in the past and as been shown to achieve very good results. \xhail, \iled  \ and \ilasp \ are crisp learners (i.e. there is no weight learning involved). \maxmargin \ was used with hand-crafted rule sets and the task was to learn weights for these rules.

\woledasp, \xhail, \iled \ and \ilasp \ are based on ASP and rely on \clingo. The implementation of \xhail \ and \iled \ is available online\footnote{\url{https://github.com/nkatzz/ORL}} and so is the most recent version of \ilasp \ (\ilaspfour)\footnote{\url{http://www.ilasp.com/download}}. \maxmargin \ is available from the \lomrf \ platform.

The original \iled \ algorithm is designed for soundness and cannot tolerate noise. To account for that in this experiment we used a noise-tolerant version (denoted by \iledhc) that learns theories in an iterative hill-climbing process: It first constructs a bottom theory (collection of bottom rules) from the mode declarations and then passes once over the data, which are presented in mini-batches, to generate a number of different theories, by generalizing the bottom theory w.r.t. each mini-batch. The theory with the best performance on the training set is retained and is subsequently further revised from each mini-batch in an additional pass over the data. The process continues, keeping the best revision at each iteration, until no improvement in performance is observed, or a max-iterations threshold is reached. In these experiments we used batch size of 100 time points with \iledhc, from which the algorithm converged in approximately 5-7 iterations over the data for \emph{meeting} and \emph{moving} and 3 iterations for \emph{rendezvous}.

\begin{figure}
	\centering
	\includegraphics[width=.49\textwidth]{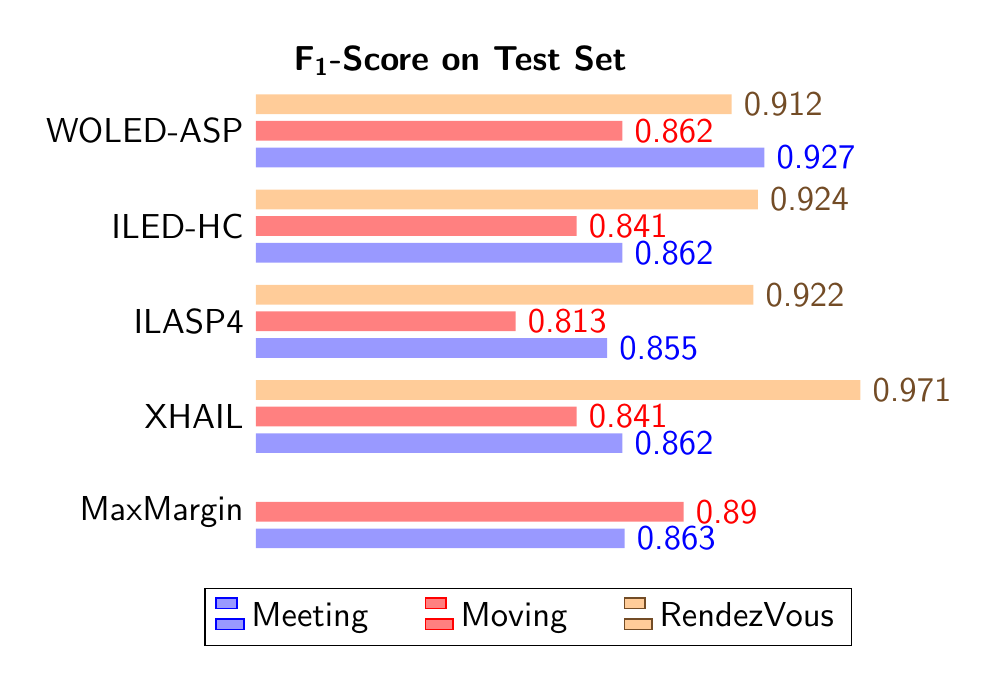}
	\label{fig:meeting}%
	\includegraphics[width=.49\textwidth]{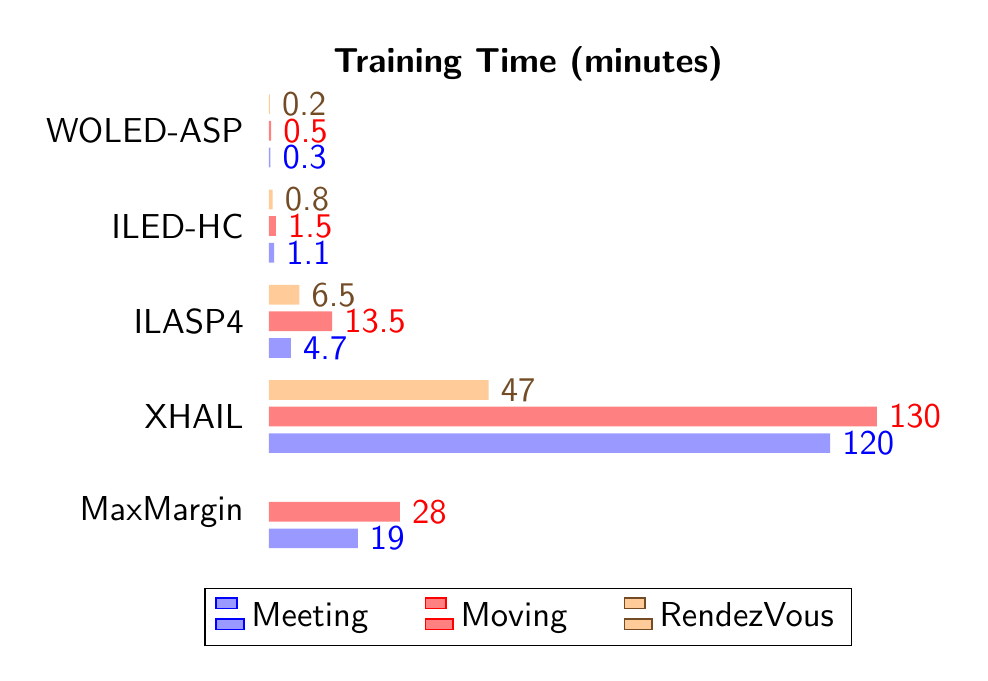}
	\caption{Comparison with batch learners.}
    \label{fig:batch-learning}
\end{figure}

We used tenfold cross-validation process for \emph{meeting} and \emph{moving} and sixfold cross-validation for \emph{rendezVous}. The results are presented in Figure \ref{fig:batch-learning} in the form of (micro-averaged) $F_1$-scores from the testing sets and average training times for each algorithm. We omit results for \maxmargin \ on \emph{rendezVous}, since the hand-crafted rules that \maxmargin \ would learn weights for are those that were used to generate the ground truth. 

\woledasp \ achieves the best $F_1$-score for \emph{meeting}, with a significant distance from the other algorithms. It also achieves the second-best $F_1$-score for \emph{moving}. In \emph{rendezVous}, \xhail \ is a clear winner w.r.t. predictive performance, while all other algorithms achieve comparable $F_1$-scores. \woledasp \ is significantly faster than \ilasp, \xhail \ and \maxmargin, while its efficiency is comparable to that of \iledhc's, which, however, is  outperformed by \woledasp. 

To appreciate the differences in performance between the algorithms being compared, it is helpful to take a look into their inner-workings. \xhail \ generalizes a bottom theory from the entirety of the training data in one go, which explains its increased training times. On the other hand, it is thanks to this strategy that \xhail \ is capable to learn better theories for \emph{rendezVous} than algorithms that process the training examples individually, thus failing to discover fragments of the \emph{rendezVous} definition, whose utility is revealed only when ``looking'' at the training data as a whole. In contrast to the \emph{rendezVous} case where the ground truth is synthetic, \xhail's learning strategy seems less useful in CAVIAR, 
where \xhail \ marginally outperforms \ilasp \ at the cost of much higher training times, it achieves identical performance to \iled \  and is outperformed by \woledasp \ on both \emph{meeting} and \emph{moving} and by \maxmargin \ on \emph{moving}. This latter observation is an indication for the merit of weight learning.  

In contrast to \xhail, \iledhc \ and \woledasp, which generate rules on demand in a data-driven fashion, \ilaspfour \ explicitly enumerates its search space. 
This yields large search spaces, 
which explains \ilasp's increased training times. 
\ilaspfour's inferior predictive performance is attributed to the learning setting, which for \ilaspfour \ follows closely the one reported in \cite{ilasp} and aims at keeping learning  tractable: first, to limit the number of irrelevant answer sets generated during learning, additional constraints on \ilaspfour's search space dictate that the learnt event patterns should only account for the ``turning points'' in a fluent's truth value. That is, initiation rules for a fluent $f$ are learnt from data points $t$, such that $f$ does not hold at $t$ and holds at $t+1$. Similarly, for termination rules, $f$ should hold at $t$ and not hold at $t+1$. Second, since such turning points in fluents' truth values are too few in the datasets, as compared to points where fluents hold/do not hold continuously, the turning point examples are weighted\footnote{The weight is translated to a cost for hypotheses that do not correctly account for these turning point examples.}, reflecting their increased importance and simulating the effect of ``over-sampling'' such examples. This allows for efficient learning with \ilaspfour \ in this domain. The downside is that learning becomes very susceptible to the even the slightest noise in the turning point examples, which explains \ilaspfour's inferior performance, as compared to the other algorithms.   

We conclude this section by pointing-out that \xhail \ and \ilaspfour \ are more general-purpose learners than the event-based algorithms they were compared to. Therefore, they may be outperformed by the latter on the particular task of learning event definitions, but on the other hand, they may be used for learning tasks that the event-based algorithms cannot.

\section{Conclusions \& Future Work}
\label{sec:conclusions}

We presented an online algorithm for learning weighted Event Calculus rules. Our system is entirely implemented in ASP and it is capable of combining temporal reasoning under uncertainty via probabilistic logical inference, with online structure and weight learning techniques. Our empirical evaluation on three datasets indicates that it compares favorably to state of the art online \& batch learners. 
Future work involves combination with semi-supervised learning, towards handling the scarcity of labeled data in streaming settings.

\section*{Acknowledgements}
This work is supported by the project entitled ``INFORE: Interactive Extreme-Scale Analytics and Forecasting'', funded by the EU’s Horizon 2020 research and innovation programme under grant agreement No 825070 and by the project SYNTELESIS ``Innovative Technologies and Applications based on the Internet of Things and Cloud Computing'' (MIS 5002521), which is funded by the Operational Programme ``Competitiveness, Entrepreneurship and Innovation'' (NSRF 2014-2020) and co-financed by Greece and the EU. We would like to thank Mark Law for his assistance in running ILASP.

\bibliographystyle{acmtrans}
\bibliography{refs}

\begin{thebibliography}{}

\bibitem[\protect\citeauthoryear{Alevizos, Skarlatidis, Artikis, and
  Paliouras}{Alevizos et~al\mbox{.}}{2017}]{DBLP:journals/corr/AlevizosSAP17}
{\sc Alevizos, E.}, {\sc Skarlatidis, A.}, {\sc Artikis, A.}, {\sc and} {\sc
  Paliouras, G.} 2017.
\newblock Probabilistic complex event recognition: {A} survey.
\newblock {\em {ACM} Comput. Surv.\/}~{\em 50,\/}~5, 71:1--71:31.

\bibitem[\protect\citeauthoryear{Artikis, Sergot, and Paliouras}{Artikis
  et~al\mbox{.}}{2015}]{artikis2015event}
{\sc Artikis, A.}, {\sc Sergot, M.}, {\sc and} {\sc Paliouras, G.} 2015.
\newblock An event calculus for event recognition.
\newblock {\em Knowledge and Data Engineering, IEEE Transactions on\/}~{\em
  27,\/}~4, 895--908.

\bibitem[\protect\citeauthoryear{Artikis, Skarlatidis, Portet, and
  Paliouras}{Artikis et~al\mbox{.}}{2012}]{artikis2012logic}
{\sc Artikis, A.}, {\sc Skarlatidis, A.}, {\sc Portet, F.}, {\sc and} {\sc
  Paliouras, G.} 2012.
\newblock Logic-based event recognition.
\newblock {\em The Knowledge Engineering Review\/}~{\em 27,\/}~04, 469--506.

\bibitem[\protect\citeauthoryear{Athakravi, Corapi, Broda, and Russo}{Athakravi
  et~al\mbox{.}}{2013}]{athakravi2013learning}
{\sc Athakravi, D.}, {\sc Corapi, D.}, {\sc Broda, K.}, {\sc and} {\sc Russo,
  A.} 2013.
\newblock Learning through hypothesis refinement using answer set programming.
\newblock In {\em Inductive Logic Programming}. Springer, 31--46.

\bibitem[\protect\citeauthoryear{Bifet, Gavald{\`a}, Holmes, and
  Pfahringer}{Bifet et~al\mbox{.}}{2018}]{bifet2018machine}
{\sc Bifet, A.}, {\sc Gavald{\`a}, R.}, {\sc Holmes, G.}, {\sc and} {\sc
  Pfahringer, B.} 2018.
\newblock {\em Machine learning for data streams: with practical examples in
  MOA}.
\newblock MIT Press.

\bibitem[\protect\citeauthoryear{Cugola and Margara}{Cugola and
  Margara}{2012}]{cugola2012processing}
{\sc Cugola, G.} {\sc and} {\sc Margara, A.} 2012.
\newblock Processing flows of information: From data stream to complex event
  processing.
\newblock {\em ACM Computing Surveys (CSUR)\/}~{\em 44,\/}~3, 15.

\bibitem[\protect\citeauthoryear{De~Raedt}{De~Raedt}{2008}]{de2008logical}
{\sc De~Raedt, L.} 2008.
\newblock {\em Logical and relational learning}.
\newblock Springer Science \& Business Media.

\bibitem[\protect\citeauthoryear{De~Raedt, Kersting, Natarajan, and
  Poole}{De~Raedt et~al\mbox{.}}{2016}]{raedt2016statistical}
{\sc De~Raedt, L.}, {\sc Kersting, K.}, {\sc Natarajan, S.}, {\sc and} {\sc
  Poole, D.} 2016.
\newblock Statistical relational artificial intelligence: Logic, probability,
  and computation.
\newblock {\em Synthesis Lectures on Artificial Intelligence and Machine
  Learning\/}~{\em 10,\/}~2, 1--189.

\bibitem[\protect\citeauthoryear{Domingos and Hulten}{Domingos and
  Hulten}{2000}]{DBLP:conf/kdd/DomingosH00}
{\sc Domingos, P.~M.} {\sc and} {\sc Hulten, G.} 2000.
\newblock Mining high-speed data streams.
\newblock In {\em {ACM} {SIGKDD}}. 71--80.

\bibitem[\protect\citeauthoryear{Duchi, Hazan, and Singer}{Duchi
  et~al\mbox{.}}{2011}]{duchi2011adaptive}
{\sc Duchi, J.}, {\sc Hazan, E.}, {\sc and} {\sc Singer, Y.} 2011.
\newblock Adaptive subgradient methods for online learning and stochastic
  optimization.
\newblock {\em Journal of Machine Learning Research\/}~{\em 12,\/}~Jul,
  2121--2159.

\bibitem[\protect\citeauthoryear{Guimar{\~{a}}es, Paes, and
  Zaverucha}{Guimar{\~{a}}es
  et~al\mbox{.}}{2019}]{DBLP:journals/ml/GuimaraesPZ19}
{\sc Guimar{\~{a}}es, V.}, {\sc Paes, A.}, {\sc and} {\sc Zaverucha, G.} 2019.
\newblock Online probabilistic theory revision from examples with proppr.
\newblock {\em Mach. Learn.\/}~{\em 108,\/}~7, 1165--1189.

\bibitem[\protect\citeauthoryear{Huynh and Mooney}{Huynh and
  Mooney}{2009}]{huynh2009max}
{\sc Huynh, T.~N.} {\sc and} {\sc Mooney, R.~J.} 2009.
\newblock Max-margin weight learning for markov logic networks.
\newblock In {\em {ECML}-2009}. Springer, 564--579.

\bibitem[\protect\citeauthoryear{Huynh and Mooney}{Huynh and
  Mooney}{2011}]{huynh2011online}
{\sc Huynh, T.~N.} {\sc and} {\sc Mooney, R.~J.} 2011.
\newblock Online max-margin weight learning for markov logic networks.
\newblock In {\em SDM}. SIAM, 642--651.

\bibitem[\protect\citeauthoryear{Katzouris}{Katzouris}{2017}]{katzourisPhd}
{\sc Katzouris, N.} 2017.
\newblock Scalable relational learning for event recognition.
\newblock {\em PhD Thesis, University of Athens\/}~{\em
  http://users.iit.demokritos.gr/~nkatz/papers/nkatz-phd.pdf}.

\bibitem[\protect\citeauthoryear{Katzouris and Artikis}{Katzouris and
  Artikis}{2020}]{kr2020}
{\sc Katzouris, N.} {\sc and} {\sc Artikis, A.} 2020.
\newblock {WOLED}: A tool for online learning weighted answer set rules for
  temporal reasoning under uncertainty.
\newblock In {\em {KR} 2020}.

\bibitem[\protect\citeauthoryear{Katzouris, Artikis, and Paliouras}{Katzouris
  et~al\mbox{.}}{2015}]{katzouris2015incremental}
{\sc Katzouris, N.}, {\sc Artikis, A.}, {\sc and} {\sc Paliouras, G.} 2015.
\newblock Incremental learning of event definitions with inductive logic
  programming.
\newblock {\em Machine Learning\/}~{\em 100,\/}~2-3, 555--585.

\bibitem[\protect\citeauthoryear{Katzouris, Artikis, and Paliouras}{Katzouris
  et~al\mbox{.}}{2016}]{DBLP:journals/tplp/KatzourisAP16}
{\sc Katzouris, N.}, {\sc Artikis, A.}, {\sc and} {\sc Paliouras, G.} 2016.
\newblock Online learning of event definitions.
\newblock {\em {TPLP}\/}~{\em 16,\/}~5-6, 817--833.

\bibitem[\protect\citeauthoryear{Katzouris, Artikis, and Paliouras}{Katzouris
  et~al\mbox{.}}{2019}]{DBLP:journals/fgcs/KatzourisAP19}
{\sc Katzouris, N.}, {\sc Artikis, A.}, {\sc and} {\sc Paliouras, G.} 2019.
\newblock Parallel online event calculus learning for complex event
  recognition.
\newblock {\em Future Gener. Comput. Syst.\/}~{\em 94}, 468--478.

\bibitem[\protect\citeauthoryear{Katzouris, Michelioudakis, Artikis, and
  Paliouras}{Katzouris et~al\mbox{.}}{2018}]{DBLP:conf/pkdd/KatzourisMAP18}
{\sc Katzouris, N.}, {\sc Michelioudakis, E.}, {\sc Artikis, A.}, {\sc and}
  {\sc Paliouras, G.} 2018.
\newblock Online learning of weighted relational rules for complex event
  recognition.
\newblock In {\em {ECML}-{PKDD} 2018}. 396--413.

\bibitem[\protect\citeauthoryear{Law, Russo, and Broda}{Law
  et~al\mbox{.}}{2015}]{ILASP_system}
{\sc Law, M.}, {\sc Russo, A.}, {\sc and} {\sc Broda, K.} 2015.
\newblock The {ILASP} system for learning answer set programs.
\newblock \url{www.ilasp.com}.

\bibitem[\protect\citeauthoryear{Law, Russo, and Broda}{Law
  et~al\mbox{.}}{2018}]{ilasp}
{\sc Law, M.}, {\sc Russo, A.}, {\sc and} {\sc Broda, K.} 2018.
\newblock Inductive learning of answer set programs from noisy examples.
\newblock Advances in Cognitive Systems.

\bibitem[\protect\citeauthoryear{Lee, Talsania, and Wang}{Lee
  et~al\mbox{.}}{2017}]{DBLP:journals/tplp/LeeTW17}
{\sc Lee, J.}, {\sc Talsania, S.}, {\sc and} {\sc Wang, Y.} 2017.
\newblock Computing {LPMLN} using {ASP} and {MLN} solvers.
\newblock {\em Theory Pract. Log. Program.\/}~{\em 17,\/}~5-6, 942--960.

\bibitem[\protect\citeauthoryear{Lee and Wang}{Lee and
  Wang}{2016}]{DBLP:conf/kr/LeeW16}
{\sc Lee, J.} {\sc and} {\sc Wang, Y.} 2016.
\newblock Weighted rules under the stable model semantics.
\newblock In {\em {KR}, 2016}.

\bibitem[\protect\citeauthoryear{Lifschitz}{Lifschitz}{2019}]{lifschitz2019answer}
{\sc Lifschitz, V.} 2019.
\newblock {\em Answer set programming}.
\newblock Springer.

\bibitem[\protect\citeauthoryear{Michelioudakis, Skarlatidis, Paliouras, and
  Artikis}{Michelioudakis et~al\mbox{.}}{2016}]{michelioudakis2016mathtt}
{\sc Michelioudakis, E.}, {\sc Skarlatidis, A.}, {\sc Paliouras, G.}, {\sc and}
  {\sc Artikis, A.} 2016.
\newblock Osla: Online structure learning using background knowledge
  axiomatization.
\newblock In {\em ECML}. Springer, 232--247.

\bibitem[\protect\citeauthoryear{Mueller}{Mueller}{2014}]{mueller2014commonsense}
{\sc Mueller, E.~T.} 2014.
\newblock {\em Commonsense reasoning: an event calculus based approach}.
\newblock Morgan Kaufmann.

\bibitem[\protect\citeauthoryear{Patroumpas, Alevizos, Artikis, Vodas, Pelekis,
  and Theodoridis}{Patroumpas et~al\mbox{.}}{2017}]{patroumpas2017online}
{\sc Patroumpas, K.}, {\sc Alevizos, E.}, {\sc Artikis, A.}, {\sc Vodas, M.},
  {\sc Pelekis, N.}, {\sc and} {\sc Theodoridis, Y.} 2017.
\newblock Online event recognition from moving vessel trajectories.
\newblock {\em GeoInformatica\/}~{\em 21,\/}~2, 389--427.

\bibitem[\protect\citeauthoryear{Ray}{Ray}{2009}]{ray2009nonmonotonic}
{\sc Ray, O.} 2009.
\newblock Nonmonotonic abductive inductive learning.
\newblock {\em Journal of Applied Logic\/}~{\em 7,\/}~3, 329--340.

\bibitem[\protect\citeauthoryear{Skarlatidis and Michelioudakis}{Skarlatidis
  and Michelioudakis}{2014}]{LoMRF}
{\sc Skarlatidis, A.} {\sc and} {\sc Michelioudakis, E.} 2014.
\newblock {Logical Markov Random Fields (LoMRF): an open-source implementation
  of Markov Logic Networks}.

\bibitem[\protect\citeauthoryear{Skarlatidis, Paliouras, Artikis, and
  Vouros}{Skarlatidis et~al\mbox{.}}{2015}]{skarlatidis2015probabilistic}
{\sc Skarlatidis, A.}, {\sc Paliouras, G.}, {\sc Artikis, A.}, {\sc and} {\sc
  Vouros, G.~A.} 2015.
\newblock Probabilistic event calculus for event recognition.
\newblock {\em ACM Transactions on Computational Logic (TOCL)\/}~{\em 16,\/}~2,
  11.

\bibitem[\protect\citeauthoryear{Tsilionis, Koutroumanis, Nikitopoulos,
  Doulkeridis, and Artikis}{Tsilionis
  et~al\mbox{.}}{2019}]{DBLP:journals/tplp/TsilionisKNDA19}
{\sc Tsilionis, E.}, {\sc Koutroumanis, N.}, {\sc Nikitopoulos, P.}, {\sc
  Doulkeridis, C.}, {\sc and} {\sc Artikis, A.} 2019.
\newblock Online event recognition from moving vehicles: Application paper.
\newblock {\em Theory Pract. Log. Program.\/}~{\em 19,\/}~5-6, 841--856.

\end{thebibliography}

\end{document}